# Deep Learning Models for Robust Facial Liveness Detection

**Oleksandr Kuznetsov***
Department of Theoretical and Applied Sciences, eCampus University, Via Isimbardi 10, Novedrate (CO), 22060, Italy,
Department of Intelligent Software Systems and Technologies, School of Computer Science and Artificial Intelligence, V. N. Karazin Kharkiv National University, Kharkiv, Ukraine
E-mail: oleksandr.kuznetsov@uniecampus.it, kuznetsov@karazin.ua
https://orcid.org/0000-0003-2331-6326

**Emanuele Frontoni**
Department of Political Sciences, Communication and International Relations, University of Macerata, Macerata, Italy, Marche Polytechnic University, Ancona, Italy
E-mail: emanuele.frontoni@unimc.it
https://orcid.org/0000-0002-8893-9244

**Luca Romeo**
Department Economics and Law, University of Macerata, Macerata, Italy
Department of Information Engineering, Marche Polytechnic University, Ancona, Italy
E-mail: luca.romeo@unimc.it
https://orcid.org/0000-0003-1707-0147

**Riccardo Rosati**
Department of Information Engineering, Marche Polytechnic University, Ancona, Italy
E-mail: r.rosati@pm.univpm.it
https://orcid.org/0000-0003-3288-638X

**Andrea Maranesi**
Department of Information Engineering, Marche Polytechnic University, Via Brecce Bianche 12, 60131 Ancona, Italy.
E-mail: andrea.maranesi99@gmail.com
https://orcid.org/0009-0007-8104-7263

**Alessandro Muscatello**
Department of Information Engineering, Marche Polytechnic University, Via Brecce Bianche 12, 60131 Ancona, Italy.
E-mail: S1107707@studenti.univpm.it
https://orcid.org/0000-0002-0927-5559

***Corresponding author: Oleksandr Kuznetsov**

## Statements and Declarations

- Competing Interests: The authors have no competing interests to declare that are relevant to the content of this article.
- Data Accessibility Statement: Public datasets (3DMAD, Replay-Attack, MSSpoof, CSMAD) were obtained from their official sources. Our custom dataset contains faces collected under public-content terms; due to consent and redistribution constraints, we do not release the raw media. We provide full preprocessing scripts, metadata, and split definitions to enable reproducibility (https://github.com/KuznetsovKarazin/liveness-detection)

**Abstract:** In the rapidly evolving landscape of digital security, biometric authentication systems, particularly facial recognition, have emerged as integral components of various security protocols. However, the reliability of these systems is compromised by sophisticated spoofing attacks, where imposters gain unauthorized access by falsifying biometric traits. Current literature reveals a concerning gap: existing liveness detection methodologies – designed to counteract these breaches – fall short against advanced spoofing tactics employing deepfakes and other artificial intelligence-driven manipulations. This study introduces a robust solution through novel deep learning models addressing the deficiencies in contemporary anti-spoofing techniques. By innovatively integrating texture analysis and reflective properties associated with genuine human traits, our models distinguish authentic presence from replicas with remarkable precision. Extensive evaluations were conducted across five diverse datasets, encompassing a wide range of attack vectors and environmental conditions. Results demonstrate substantial advancement over existing systems, with our best model (AttackNet V2.2) achieving 99.9% average accuracy when trained on combined data. Moreover, our research unveils critical insights into the behavioral patterns of impostor attacks, contributing to a more nuanced understanding of their evolving nature. The implications are profound: our models do not merely fortify the authentication processes but also instill confidence in biometric systems across various sectors reliant on secure access. By achieving near-perfect accuracy while maintaining computational efficiency, this research illuminates the path forward in combating identity fraud, ensuring data privacy, and safeguarding sensitive information in the digital realm, thus providing an impetus for further innovations in cybersecurity frameworks.

# 1. Introduction

## 1.1. Background of the Problem

Liveness detection is an essential component in biometric authentication systems, serving as a crucial line of defense against spoofing attacks [1]. Its primary objective is to discern real, live biometric characteristics from fraudulent copies or recreations [2]. This detection plays a key role in various applications such as facial recognition, fingerprint scanning, and iris recognition among others [3].

Historically, simple biometric systems relied on static information, like a stored image of a face or fingerprint for verification [4]. However, these systems were susceptible to spoofing by presenting fake biometric traits, a photo in the case of facial recognition, or a cast in the case of fingerprints, which has posed a substantial threat to the security and reliability of these systems.

The escalating complexity of these threats, coupled with the rising ubiquity of biometric systems in everyday life, from smartphones to international airport security, necessitated the development of dynamic liveness detection techniques [3]. Modern liveness detection systems are designed to measure signs of life, including but not limited to, minor involuntary facial movements, unique patterns in the way a person speaks, or blood flow beneath the skin of a finger.

Despite the impressive advancements in liveness detection technologies over the past few years, significant challenges persist. Sophisticated attacks continue to evolve, utilizing emerging technologies such as high-definition printing, 3D modeling, deepfakes, and even synthetic biometric traits [4].

These ongoing challenges emphasize the need for continual research, refinement, and enhancement of liveness detection methods. Consequently, our work strives to address these contemporary threats, providing innovative solutions that promise to bolster the effectiveness and reliability of biometric authentication systems.

In the realm of facial biometrics, the challenge of liveness detection, specifically, the detection of spoofing attacks, becomes exponentially complex due to the three-dimensional and dynamic nature of human faces, coupled with the plethora of ways in which a facial spoof can be orchestrated.

Traditional methods of face liveness detection have included texture analysis, wherein certain patterns or artifacts left behind by fake replicas are identified, and motion-based techniques that bank on certain predictable or repeatable human facial movements [5]. However, these methods have shortcomings. Texture analysis techniques often struggle with high-quality fakes, while motion-based approaches can be bypassed through sophisticated replicas and user behavior imitation.

The advent of deep learning has revolutionized liveness detection techniques by bringing in the capability to model and learn high-dimensional patterns, thus mitigating the limitations of conventional methods. Techniques based on Convolutional Neural Networks (CNNs) and Recurrent Neural Networks (RNNs) have shown great promise, with the ability to learn and discern intricate patterns from large volumes of data [6]. Furthermore, generative models such as GANs (Generative Adversarial Networks) have been utilized to create synthetic liveness features, improving the robustness of detection models.

Yet, deep learning methods also have their limitations. They demand significant volumes of data for training, and they need to be periodically updated to adapt to the emerging threats. Also, they may sometimes struggle with generalization, especially when encountering unseen types of spoofing attacks.

Specifically, current challenges include: (1) severe performance degradation in cross-dataset scenarios, with accuracy dropping from >95% to <50% when models encounter unseen attack types; (2) inability to generalize across different acquisition devices and environmental conditions; (3) vulnerability to high-quality silicone masks and digital replay attacks that closely mimic genuine biometric traits. These gaps necessitate the development of more robust architectures capable of learning universal spoofing patterns rather than dataset-specific artifacts.

### 1.2. Relevance of the Study

In today's digital age, face recognition systems have become integral to a wide array of applications, ranging from smartphone unlocking, surveillance, social media, and banking, among others. As these systems gain prominence, the likelihood of spoofing attacks increases, making liveness detection an indispensable component to ensure the security and integrity of these systems [7, 8].

However, due to the increasing sophistication of spoofing attacks and the limitations of current liveness detection techniques, the need for a robust, adaptable, and reliable solution is more pressing than ever. Our research investigates the use of advanced deep learning methods for face liveness detection, addressing the aforementioned challenges while expanding the boundaries of the current state-of-the-art. Our efforts in this domain are not only crucial for enhancing security but also have far-reaching implications for fostering trust and reliability in face recognition technologies, which have become an integral part of our everyday lives.

### 1.3. Objectives of the Study

Through this research, we aim to explore, understand, and overcome the challenges that face liveness detection in the context of face recognition systems. Leveraging various spoofing attack scenarios and corresponding datasets, we set forth several objectives that are geared towards advancing the field of liveness detection.

**Objective 1:** Evaluate and learn from various spoofing attacks, with each presenting a unique threat scenario. To this end, we use several datasets:

- Custom Silicone Mask Attack (CSMAD) dataset, providing us insights into high-quality silicone mask attacks manufactured by a professional special effects company [9].
- The 3D Mask Attack Database (3DMAD), exposing us to biometric 3D data captured via Kinect [10].
- Multispectral-Spoof Database (MSSPOOF), which gives us access to both visible (VIS) and near-infrared (NIR) spectrum images and print spoofing attacks [11].
- Replay-Attack Database, comprising videos of photo and video attack attempts under various lighting conditions [12].
- Our custom dataset made up of bona fide and attacker images extracted from videos taken via smartphone and laptop webcam.

**Objective 2**: Use a modern face detector to extract, align, and crop faces across all datasets; Standardize resolution to 256×256 and apply quality control; Create subject-disjoint train/validation/test splits and document the split statistics per dataset; Benchmark strong existing models (e.g., LivenessNet, AttackNet variants) and report PAD metrics (APCER, BPCER, ACER, EER, HTER) together with ROC-AUC/PR-AUC; Analyze error patterns by attack type (print, replay, mask) to identify failure modes that motivate a new architecture.

**Objective 3**: Design a compact CNN for PAD with standard best practices; Enhance robustness using state-of-the-art training techniques: strong but realistic augmentation (geometric/photometric, blur, noise), L2 weight decay, dropout, label smoothing, learning-rate scheduling, early stopping, and class balancing; Improve generalization via rigorous evaluation protocols: zero-shot cross-dataset testing and combined-dataset training to mitigate domain shift; select thresholds on validation only and keep them fixed for test.

Through these objectives, we aim to make significant contributions to liveness detection in facial biometrics, with the potential of setting a new benchmark for the field. More than just academic accomplishment, however, our endeavor is targeted at fostering trust and security in face recognition systems – a pivotal step towards a secure digital future.

### 1.4. Overview of the Article

This article is structured to provide comprehensive insight into our exploration of liveness detection in face recognition systems, using deep learning techniques. Each section of the paper is crafted to guide the reader seamlessly through our research journey, thereby facilitating a complete understanding of our study and its significance.

The paper begins with a Literature Review (Section 2), in which we delve into the existing body of knowledge, examining previous studies, and identifying the gaps that our research aims to fill. This exploration of the academic landscape enables us to position our work within the broader context of liveness detection research.

In the Methodology section (Section 3), we discuss the data we used in our research, detailing the diverse datasets that represent various spoofing attack scenarios. We also detail the techniques used for face extraction and deep learning models that have been tested, along with the description of our novel CNN architecture.

Section 4, titled Results, presents the findings from our experiments, illustrating how our proposed model performs against each spoofing attack scenario. This section discusses the strengths and potential weaknesses of our model, validated through our diverse datasets.

Next, the Discussion section (Section 5) provides a platform for interpreting our results in the broader context of liveness detection. We compare our findings with those from previous literature, discussing the implications and potential applications of our research.

The final Conclusion section (Section 6) offers a summary of our key findings and their significance. It also elucidates the potential impact of our work and suggests areas for future research. This section underlines the broader goals of our study - reinforcing the security of face recognition systems and fostering trust in these technologies.

Throughout the paper, we have made an effort to articulate our processes, findings, and interpretations in an accessible yet rigorous academic style. The goal is to allow readers from various backgrounds to engage with our work, encouraging the exchange of ideas and promoting further advancements in the field.

We delimit scope to presentation attacks in RGB face PAD. Voice and iris liveness are discussed in Related Work to contextualize artifacts and embedded constraints.

## 2. Literature Review

In the field of face anti-spoofing, there is a rich tapestry of research that our study builds upon. However, existing works leave gaps that provide opportunities for further investigation and improvement.

Chingovska et al. [13] probed the potential of Local Binary Patterns (LBP) texture features on three types of face spoofing attacks. Their findings suggest moderate discriminability when tested against a variety of attack types, with an achieved Half Total Error Rate (HTER) of around 15%. While LBPs have shown some promise, their modest performance underlines the need for more effective techniques.

Erdogmus and Marcel [14] explored the spoofing potential of subject-specific 3D facial masks for 2D face recognition systems. Their research revealed a high vulnerability to these types of attacks, achieving an HTER of about 20% using LBP-based countermeasures. This reinforces the need for more effective solutions that can address such sophisticated spoofing techniques.

Bhattacharjee et al. [15] examined the vulnerability of CNN based face-recognition systems against spoofing attacks using custom-made silicone masks. They demonstrated that these attacks present a significant threat, with the vulnerability of each system being at least ten times higher than its false match rate. Their proposed solution, a simple presentation attack detection method using a low-cost thermal camera, is not based on deep learning techniques, underscoring the gap that exists in the effective application of deep learning methods against such attacks.

Chingovska et al. [16] reported on the significant security risk that spoofing attacks pose for face recognition systems across the visual spectrum. They achieved an HTER of around 8-9% using various methods, which could be reduced to 5-7% with multispectral processing. However, these rates are still far from ideal, signifying the need for more robust countermeasures.

Alotaibi and Mahmood [17] explored the use of specialized deep convolution neural networks in detecting face spoofing attacks. Their method, which processed a single frame of sequential frames, resulted in an HTER of about 4% using the Replay Attack dataset. While this work showcased the potential of deep learning in spoofing detection, the relatively high HTER illustrates that there is still room for improvement.

Sun et al. [18] undertook a comprehensive investigation of different supervision schemes in face spoofing detection using depth-based Fully Convolutional Networks (FCNs). Their proposal of a Spatial Aggregation of Pixel-level Local Classifiers (SAPLC) approach yielded competitive performance on several datasets. However, the cross-database testing on the Replay Attack dataset resulted in a significantly high HTER of around 30%, indicating that transferability across different datasets is still a major issue in this field.

Kotwal and Marcel [19] proposed a novel patch pooling mechanism integrated with a pre-trained CNN for detecting 3D mask presentation attacks in near-infrared (NIR) imaging. Their method demonstrated efficacy on mask attacks in the NIR channel, achieving near-perfect results on the WMCA dataset and outperforming the existing benchmark on the MLFP dataset. Despite their promising results, their study was limited to NIR imaging and 3D mask attacks, leaving the applicability of their method to other attack types and imaging modalities unexplored.

Mallat and Dugelay [20] introduced a novel type of attack on thermal face recognition systems and demonstrated the vulnerability of these systems to such attacks. This study indicates a critical need for further research into the security of face recognition systems, especially with the development of increasingly sophisticated attack techniques.

Wang et al. [21] proposed a novel silicone mask face anti-spoofing detection method based on visual saliency and facial motion characteristics. Despite its superiority over existing methods in public and self-built datasets, the resulting HTER of around 9% indicates that there are still gaps in the effectiveness of current anti-spoofing methods against sophisticated attacks such as silicone masks.

Finally, Arora et al. [22] proposed a robust framework for face spoofing detection, relying on the extraction of features from faces using pre-trained convolutional autoencoders. Their study achieved an HTER of about 4% on several datasets. However, their cross-database testing yielded an HTER of around 40%, emphasizing that achieving a robust model capable of generalizing across different datasets remains a challenge.

Recent studies report strong within-dataset accuracy with compact backbones and transfer learning. LwFLNeT, a dual-stream lightweight CNN with parallel dropout, attains very low HTER on 3DMAD, NUAA, and Replay-Attack under within-dataset protocols, but degrades under cross-dataset transfer to 3DMAD (Shinde et al., 2025 [23]). A systematic evaluation of pre-trained CNNs shows DenseNet-based models with ACER near 1–2% on NUAA/Replay and competitive cross-dataset HTER between NUAA and Replay. MobileNetV2 achieves real-time inference while retaining accuracy (Khairnar et al., 2025 [24]). These results confirm a common pattern: excellent in-domain performance and a persistent gap under domain shift.

Adaptive Lipschitz-bound regularization has been proposed to reduce overfitting by constraining layer-wise spectral norms. The method adapts the constraint during training and improves the train–validation gap across several benchmarks (Chacón-Chamorro et al., 2026 [25]). Although not specific to PAD, this line of work supports our choice to prioritize stability-oriented regularization over excessive architectural complexity.

Lightweight design is also explored through feature smoothing and sparse skip connections to reduce parameters and redundancy while preserving accuracy (Li et al., 2023 [26]). These ideas are relevant for PAD deployments on edge devices, where latency and memory budgets are strict.

Pupillary light reflex (PLR) offers a stimulus-driven liveness cue. However, recent results show high error rates on Replay-Attack and CASIA-SURF under operational protocols, indicating limited reliability without careful control of capture conditions (Prasad et al., 2023 [27]). We treat PLR as complementary rather than a replacement for appearance-based PAD.

Voice liveness is advancing via spectral transforms and artifact-aware features. One line of work detects “pop-noise” as a low-frequency cue of live speech using Constant-Q or Morse-wavelet representations, with competitive accuracy on ASVspoof and POCO datasets (Gupta and Patil, 2024 [28]; Eyidoğan et al., 2025 [29]). These approaches highlight the value of physically grounded artifacts; the concept informs, but does not directly transfer to face PAD.

For iris PAD, texture-feature fusion (LBP+GLCM) has been optimized for embedded systems with real-time feasibility (Tran et al., 2024 [30]). Other works exploit handcrafted energy features with classical ensembles and report high within-dataset accuracy (Khade et al., 2023 [31]). These results emphasize the practicality of lightweight pipelines in ocular biometrics.

Deepfake detection targets identity-preserving forgeries rather than presentation attacks. Hybrid CNN–Transformer models achieve very high AUC on DeepForensics and CelebDF (Siddiqui et al., 2025 [32]). We do not evaluate deepfakes in this study; we cite them to delimit scope and terminology.

The recent literature strengthens three points. First, within-dataset PAD can approach near-zero error, yet cross-dataset transfer remains challenging. Second, efficiency-oriented designs are increasingly preferred for deployment. Third, theory-driven regularization may complement data-centric strategies to improve robustness. Our experimental design and discussion align with these trends.

## 3. Methodology

### 3.1. Data

Our research employs several datasets representing different spoofing attack scenarios.
We use the full public versions of 3DMAD, Replay-Attack, MSSpoof, and CSMAD, except for frames removed by our quality filter (blur/occlusion thresholds). The data are challenging and cover print, replay, and mask presentation attacks across varying capture conditions and sensors. Our experiments follow subject-disjoint splits and a zero-shot cross-dataset protocol to stress domain shift.

### 3.1.1. Custom Silicone Mask Attack Dataset (CSMAD)

The Custom Silicone Mask Attack Dataset was collected at the Idiap Research Institute and is specifically designed for face presentation attack detection experiments, primarily focusing on presentation attacks mounted using a custom-made silicone mask of the person being attacked [9, 15].

#### Data Collection

The CSMAD comprises face-biometric data from 14 subjects. Each subject has performed three roles: as targets, attackers, and bona-fide clients. Six of these subjects (identified as A to F) served as targets, implying that their facial data was used to construct corresponding custom-made flexible silicone masks.
The masks were manufactured by Nimba Creations Ltd., a company specializing in special effects. These high-quality masks present a sophisticated spoofing scenario, challenging our liveness detection system.
The dataset contains both bona fide and attack presentations, increasing variability and better representing real-world conditions. The attack presentations were created by having different subjects wear the six masks. Four distinct lighting conditions were used during the data collection process, including:

1. Fluorescent ceiling light only
2. Halogen lamp illuminating from the left of the subject only
3. Halogen lamp illuminating from the right only
4. Both halogen lamps illuminating from both sides simultaneously

All presentations were captured against a uniform green background.

#### Dataset Structure

The CSMAD is structured into three subdirectories: 'attack', 'bonafide', and 'protocols'. The 'attack' and 'bonafide' directories contain videos and still images for attack and bona fide presentations, respectively, while 'protocols' include text files specifying the experimental protocol for vulnerability analysis of face recognition (FR) systems.

- The 'bonafide' directory comprises 87 videos and 17 still images, captured using a Nikon Coolpix digital camera.
- The 'attack' directory consists of 159 videos, subdivided into two categories: 'WEAR' with 108 videos, and 'STAND' with 51 videos. 'WEAR' contains videos where the attacker is wearing the mask of the target, while 'STAND' features videos where the target's mask is mounted on a stand for the attack.

This dataset provides us with a rich and varied set of data for training and testing our deep learning models, enabling us to better understand the dynamics of silicone mask spoofing attacks. The variety of subjects, masks, and lighting conditions represented in the dataset ensures our model's robustness and adaptability to various real-world conditions.
The Table 1 summarizes key aspects of the CSMAD.

Table 1: Key aspects of Custom Silicone Mask Attack Dataset

| Attribute | Details |
|---|---|
| Source | Idiap Research Institute |
| Number of Subjects | 14 |
| Number of Masks | 6 |
| Roles Performed by Subjects | Targets, Attackers, Bona-fide Clients |
| Number of Videos ('bonafide') | 87 |
| Number of Images ('bonafide') | 17 |
| Number of Videos ('attack') | 159 (108 'WEAR', 51 'STAND') |
| Lighting Conditions | 4 |

### 3.1.2. The 3D Mask Attack Database (3DMAD)

The second dataset used in our study is the 3D Mask Attack Database, another face biometric spoofing database designed for testing the robustness of face recognition systems against spoofing attacks [10, 14].

**Data Collection**

The 3DMAD contains 76,500 frames of 17 subjects, all recorded using a Kinect device. Each frame features a depth image, the corresponding RGB image, and manually annotated eye positions. The data collection process was spread across three sessions for each subject, with each session comprising five videos of 300 frames. These recordings were performed under controlled conditions, ensuring a frontal view and neutral expression.

The first two sessions contain bona fide access samples, with a time delay of approximately two weeks between the acquisitions to introduce temporal variability. The third session was dedicated to capturing 3D mask attacks, which were conducted by a single operator.

To augment the usability of the dataset, the eye positions in each video were manually labeled for every 60 frames and linearly interpolated for the rest.

The 3D masks used in this dataset were created by "ThatsMyFace.com", using 1 frontal and 2 profile images of the subjects. The database also includes these face images used for mask generation and paper-cut masks produced using the same images.

**Dataset Structure**

The 3DMAD contains a wealth of information critical to understanding and countering 3D mask spoofing attacks. Each frame consists of:

1. A depth image (640x480 pixels – 1x11 bits)
2. The corresponding RGB image (640x480 pixels – 3x8 bits)
3. Manually annotated eye positions (with respect to the RGB image)

The table 2 summarizes the key aspects of the 3DMAD.

Table 2: Key aspects of 3D Mask Attack Database

| **Attribute** | **Details** |
|---|---|
| Source | Idiap Research Institute |
| Number of Subjects | 17 |
| Frames per Subject | 4500 (5 videos of 300 frames each, over 3 sessions) |
| Data per Frame | Depth Image, RGB Image, Annotated Eye Positions |
| Mask Creator | "ThatsMyFace.com" |
| Sessions | 3 (2 Bona fide, 1 Spoofing) |

The comprehensive and diverse data provided by 3DMAD offers an extensive platform to train and test our models, ensuring their efficacy in detecting 3D mask spoofing attacks, an emerging threat in face recognition systems. This dataset, coupled with our previously described CSMAD, forms a substantial basis for understanding and mitigating a range of spoofing attacks.

### 3.1.3. Multispectral-Spoof Database (MSSPOOF)

The third dataset used in our study is the Multispectral-Spoof Database, a spoofing attack database built at the Idiap Research Institute. This dataset specializes in recording face images under visible (VIS) and near-infrared (NIR) spectra for both real accesses and spoofing attacks [11, 16].

**Data Collection**

The MSSPOOF comprises data from 21 subjects, recorded using a uEye camera with a resolution of 1280x1024 pixels. For NIR images, a NIR filter of 800nm was mounted on the camera.

In the real access recordings, five images each in VIS and NIR were captured for each subject under seven different environmental conditions, totaling to 70 real access images per client.

For the spoofing attacks, three best quality images each from VIS and NIR spectra were selected from the original database for each client and printed on paper using a black & white printer with a resolution of 600dpi. Four spoofing attacks were recorded for each printed image under three lighting conditions, both in VIS and NIR spectra, giving a total of 144 spoofing attacks per client.

**Dataset Structure**

The recorded images are divided into training, development, and testing subsets with non-overlapping clients. The training subset includes 9 clients, while the development and test subsets contain 6 clients each. The enrollment set consists of 10 real access images per client, five each from VIS and NIR spectra.

Each sample in the MSSPOOF provides manual annotations for the facial region, given with the (x,y) coordinates of 16 key points on the face.

The table 3 summarizes the key aspects of the MSSPOOF.

Table 3: Key aspects of Multispectral-Spoof Database

| **Attribute** | **Details** |
|---|---|
| Source | Idiap Research Institute |
| Number of Subjects | 21 |
| Real Access Images per Client | 70 (35 VIS, 35 NIR) |
| Spoofing Attacks per Client | 144 |
| Spectra | VIS and NIR |
| Subsets | Training, Development, Test, Enrollment |

The MSSPOOF offers an extensive platform to test our models' performance against both VIS and NIR spoofing attacks. This dataset's multispectral nature ensures our model's robustness across various light spectra, enhancing the model's adaptability to different real-world conditions.

### 3.1.4. Replay-Attack Database

The fourth dataset in our research is the Replay-Attack Database, designed specifically for face spoofing studies. It comprises 1300 video clips of photo and video attack attempts performed on 50 clients under varying lighting conditions [12, 13].

**Data Collection**

The Replay-Attack Database splits its data into four subsets: training, development, test, and enrollment. Each of these subsets contains unique clients, ensuring no overlap across subsets.

The videos in the database capture real client access attempts or spoofing attacks using photo or video playbacks. All videos are recorded using a built-in webcam on a Macbook laptop, yielding color videos with a resolution of 320x240 pixels. The videos are saved in ".mov" format, with a frame rate of around 25 Hz.

The dataset takes into account two different lighting conditions during data collection. The controlled condition has the office light turned on, with blinds down, and a homogeneous background, while the adverse condition has blinds up, more complex background, and office lights turned off.

The spoofing attacks employ high-resolution photos and videos of each client taken under the same conditions as their authentication sessions. The attack videos utilize various methods and devices, including mobile attacks using an iPhone 3GS screen, high-resolution screen attacks using a first-generation iPad, and hard-copy print attacks produced on a color laser printer.

**Dataset Structure**

The Replay-Attack Database's distribution across the four subsets is as follows:

- Training set: Contains 60 real accesses and 300 attacks.
- Development set: Contains 60 real accesses and 300 attacks.
- Test set: Contains 80 real accesses and 400 attacks.
- Enrollment set: Contains 100 real-accesses under different lighting conditions, intended to study the baseline performance of face recognition systems.

In addition to the video data, the database provides face locations automatically annotated by a cascade of classifiers based on Modified Census Transform (MCT).

Here is a summary of the Replay-Attack Database's key aspects:

The table 4 summarizes the key aspects of the Replay-Attack Database.

Table 4: Key aspects of Replay-Attack Database

| **Attribute** | **Details** |
|---|---|
| Source | Idiap Research Institute |
| Number of Clients | 50 |
| Real Access Attempts per Subset | 60 (Training), 60 (Development), 80 (Test), 100 (Enrollment) |
| Attack Attempts per Subset | 300 (Training), 300 (Development), 400 (Test) |
| Video Resolution | 320x240 pixels |
| Subsets | Training, Development, Test, Enrollment |

The Replay-Attack Database's comprehensive video data under varying lighting conditions and diverse attack methods adds substantial depth to our study. Its unique distribution structure facilitates a structured evaluation of our models across various stages of training and testing.

### 3.1.5. Our Custom Dataset

The final dataset used in our research is a custom-made dataset, specifically crafted to simulate real-world conditions of potential spoofing attacks.

**Data Collection**

This dataset is divided into two sets of images: bona fide and attackers. The bona fide set consists of images extracted from videos showcasing actual individuals. These videos are either taken with a smartphone or downloaded from the internet.

The attacker set contains images extracted from videos that capture the playback of the bona fide videos on a smartphone screen or vice versa. The videos are recorded using a laptop webcam, simulating a typical scenario where a fraudster may try to deceive a facial recognition system using a playback video.

The primary source of our videos is YouTube, a platform offering diverse real-world conditions. This approach allows us to capture a wide range of scenarios, including different lighting conditions, angles, and individual characteristics, providing a more comprehensive and challenging dataset for training and testing our model.

**Dataset Structure**

The dataset comprises 4656 images, evenly distributed between bona fide and attacker classes, ensuring a balanced dataset for accurate model evaluation. The training and validation split is 48/52, with 2238 images allocated for training and 2418 for validation. Both the training and validation sets maintain a 50/50 class distribution to prevent bias in our model.

We extracted frames from each video and performed undersampling to balance the classes in both the training and testing datasets. This step is essential to prevent our model from overfitting to a particular class and improve its generalization capability.

The dataset is further organized into 84 videos as follows:

- 40 for training: 15 for bona fide and 16 for attackers.
- 25 for testing: 25 for bona fide and 28 for attackers.

The table 5 summarizes the key aspects of our custom dataset.

Table 5: Key aspects of Our Custom Dataset

| **Attribute** | **Details** |
|---|---|
| Total Images | 4656 |
| Class Distribution | 50/50 (bona fide/attackers) |
| Training/Validation Split | 48/52 |
| Training Images | 2238 |
| Validation Images | 2418 |
| Training Class Distribution | 50/50 (bona fide/attackers) |
| Validation Class Distribution | 50/50 (bona fide/attackers) |

Our custom dataset's unique composition makes it an invaluable resource in this study. It mimics realistic attack scenarios and ensures a diverse and balanced collection of data, helping our model learn and generalize effectively. This comprehensive dataset will aid in accomplishing our research goals, which will be discussed in more detail in the subsequent sections.

## 3.2. Liveness Detection Models

### 3.2.1 LivenessNet: Our Initial Deep Learning Model

Our research study commences with the Liveness Detection Model, initially developed by Adrian Rosebrock in 2019 [33]. The model's architecture is built upon a CNN using the Keras API from TensorFlow [34, 35].

**Architecture**

The Liveness Detection Model comprises sequential layers, forming a stack in which each layer passes its output to the subsequent layer. This architecture is designed to distinguish between authentic human faces and imitations. It incorporates an assortment of Keras layers, including Convolutional 2D (Conv2D), MaxPooling2D, Dropout, Flatten, Dense, and BatchNormalization layers.

The architecture starts with the input shape of the images being defined. The model uses a 'channels last' format where the depth is the last dimension in the input shape tuple (height, width, depth). However, if the 'channels first' format is used, the input shape and channels dimension are updated accordingly.

**Convolutional and Max Pooling Layers**

The first part of the network consists of two blocks, each with two Conv2D layers followed by a MaxPooling2D layer. Conv2D layers apply 2D convolution over the input signal, a technique integral to CNNs that provides them with their ability to learn image features. The Conv2D layers in both blocks of the model use 16 and 32 filters respectively with a kernel size of 3x3.

Each Conv2D layer is followed by an Activation layer with a Rectified Linear Unit (ReLU) function, which adds non-linearity to the network. BatchNormalization follows the activation function, improving the speed, performance, and stability of the network.

The Conv2D layers are followed by a MaxPooling2D layer with a pool size of 2x2. MaxPooling reduces the spatial dimensions (width, height) of the input, controlling overfitting by providing an abstracted form of the representation.

Each block is completed with a Dropout layer at a rate of 0.25, a technique that helps prevent overfitting by randomly ignoring a fraction of the input nodes.

**Flattening and Dense Layers**

Post the convolutional blocks, the model uses a Flatten layer to transform the 2D matrix to a 1D vector, enabling it to be processed by Dense layers.

Subsequently, the architecture employs a Dense layer with 64 neurons, followed by another ReLU activation function. A BatchNormalization layer follows this, and a Dropout layer with a rate of 0.5 is applied for regularization, once again to avoid overfitting.

The final layer of the model is another Dense layer with a softmax activation function. The softmax function outputs a vector that represents the probability distributions of a list of potential outcomes. It's a generalization of the sigmoid function, aptly suitable for multiclass classification tasks.

Figure 1 illustrates the LivenessNet architecture, a base model for our liveness detection task. The network architecture involves a sequence of Convolutional (Conv2D), Activation (ReLU), Batch Normalization, and Max Pooling layers, followed by Dense layers towards the end. The presence of Dropout layers ensures the model's robustness against overfitting. It uses a 'channels last' image data format and starts with a feature map of 16 filters, which is extended to 32 filters in the subsequent layers.

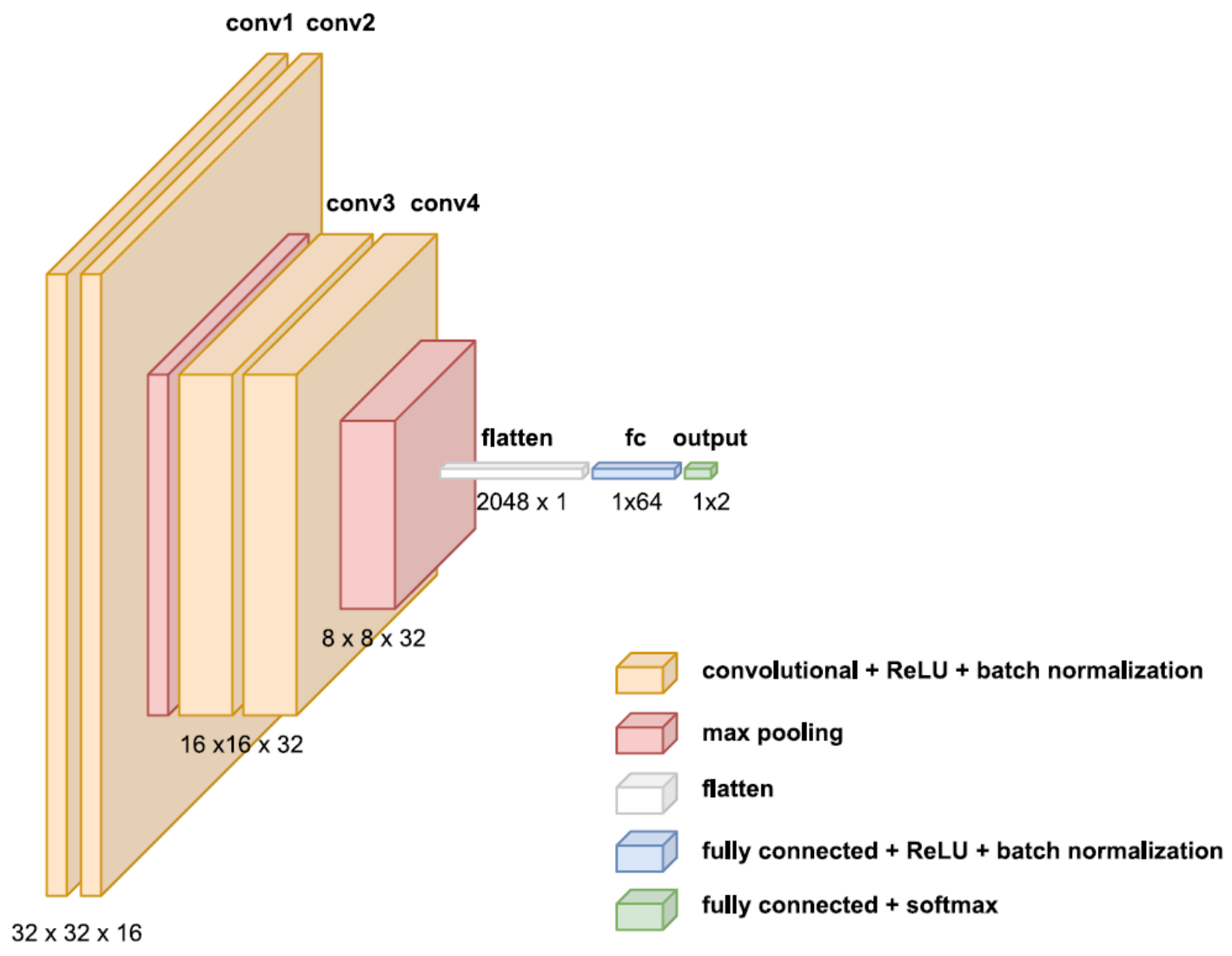

Figure 1: LivenessNet Architecture

In summary, the Liveness Detection Model is a simple yet effective network for face liveness detection tasks. The architecture focuses on maintaining a balance between model complexity and performance while ensuring that the network can extract useful features and learn from the data effectively without significant overfitting.

### 3.2.2 AttackNet v1 Architecture

To improve upon the performance of the LivenessNet architecture, we enriched its complexity while maintaining the network's efficiency. Inspired by the effectiveness of skip connections and residual blocks in combating the infamous "vanishing gradient" problem and overfitting [36], we developed a more sophisticated architecture termed as "AttackNet v1."

The salient characteristic of the AttackNet v1 architecture is the addition of an extra convolutional layer for each convolutional step, and the inclusion of skip connections, following the design cues from the widely adopted ResNet architecture [36]. The use of skip connections, or shortcut connections, has been acknowledged as a sound practice to prevent gradient explosion or vanishing, making it particularly suitable for the task of spoof detection.

The architecture's objective is threefold: to obtain impressive results on diverse datasets, to enable real-time detection with quick inference time, and to ensure feasible implementation on cost-effective hardware.

Each convolutional step of AttackNet v1 now includes one additional Conv2D layer. The purpose of this layer is to extract a higher level of features from the input data, thereby improving the model's learning capability.

Subsequently, following the prototype of ResNet [36], skip connections are applied to these layers. Rather than just linking the input of a layer to its output (as seen in traditional ResNet), these connections concatenate the outputs of Conv2D layers. Concatenation, in this case, involves merging the feature maps produced by two layers, thus preserving more information for the network to learn from. This approach provides a more expressive feature representation and enhances the network's capacity to learn complex patterns.

Figure 2 presents the architecture of AttackNet v1, a modification and extension of the LivenessNet model. The architecture is enhanced with additional convolutional layers and introduced skip connections, borrowing from the ResNet style, to prevent vanishing gradient problems. This forms the basis for further development in our sequence of models.

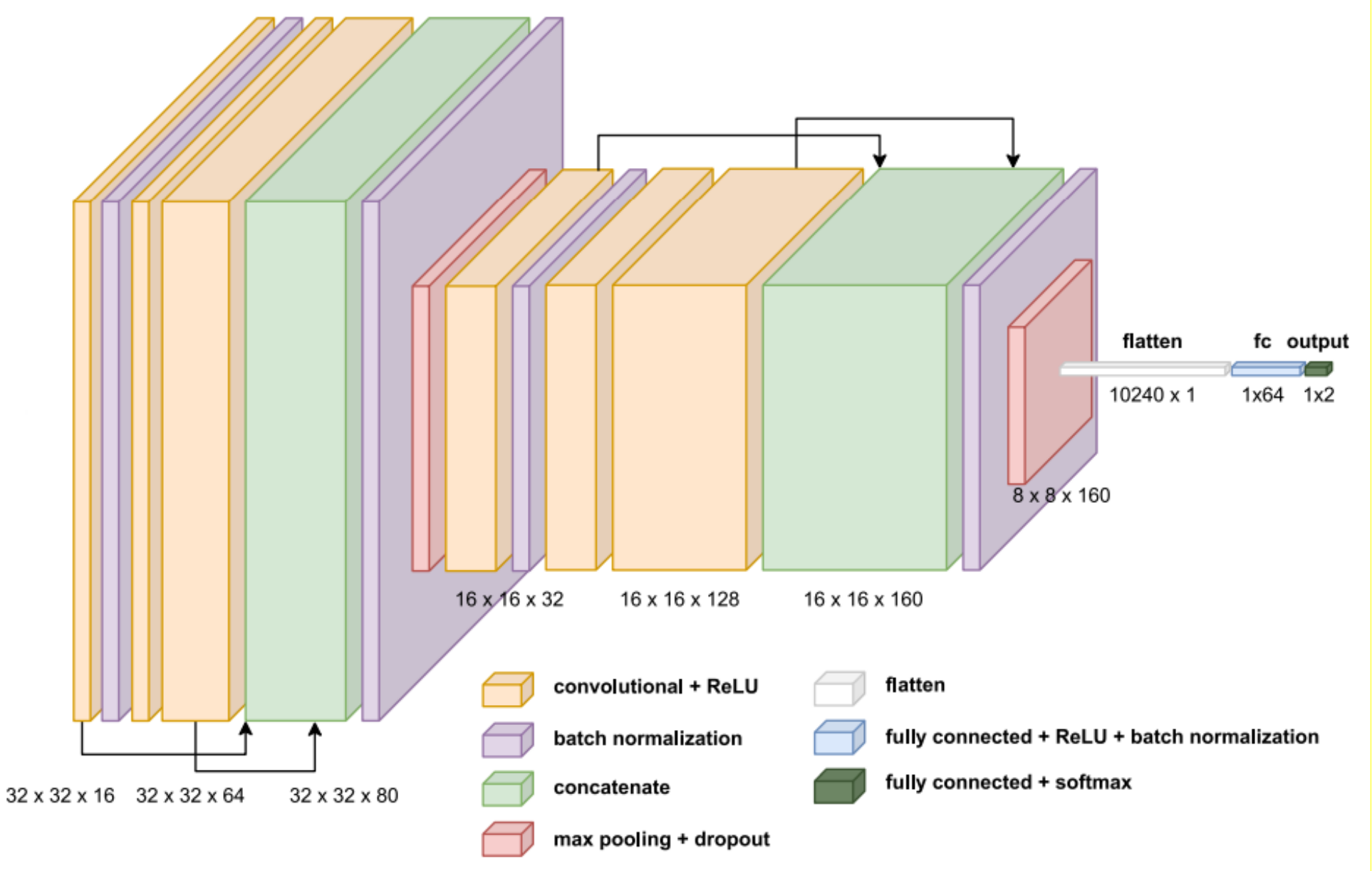

Figure 2: AttackNet v1 Architecture

In summary, the enhanced AttackNet v1 architecture makes use of additional convolutional layers and skip connections to ensure an efficient learning process. This architecture is designed not only to provide a more robust performance in face liveness detection tasks but also to ensure rapid inference and affordability, crucial for real-time application and implementation on low-cost hardware.

### 3.2.3 AttackNet v2.1 Architecture

In our quest for optimal performance, we perceived further scope for refinement in our model architecture. Given the integration of Batch Normalization in our network, we are equipped with normalized inputs for subsequent layers. This presented an opportunity to experiment with different activation functions, such as Leaky ReLU (Rectified Linear Unit) and Hyperbolic Tangent (Tanh), to potentially mitigate the loss of information in scenarios of negative input values.

In the resulting architecture, denoted as AttackNet v2.1, we adopted the LeakyReLU activation function throughout the convolutional layers to address the "dying ReLU" problem, while employing Hyperbolic Tangent (Tanh) activation for the fully-connected layer. LeakyReLU allows small negative values when the input is less than zero ($\alpha$=0.2 in our implementation), thereby preserving information about negative inputs.

Hyperbolic Tangent, on the other hand, is a mathematical function known for its usage in neural networks as an activation function. It can map any real-valued number to the range between -1 and 1, thereby ensuring that the variance of the output values remains manageable throughout the network's layers. This makes the network less likely to fall prey to the "vanishing gradients" problem, a common occurrence in large neural networks where gradients are squashed through multiple layers.

Figure 3 depicts the architecture of the AttackNet v2.1 model, which introduces new activation functions, namely LeakyReLU and Hyperbolic Tangent (Tanh), in its architecture. The LeakyReLU function helps retain some information even for negative input values, unlike the standard ReLU activation. The Tanh activation function enables the model to manage a broader range of input values, from -1 to 1.

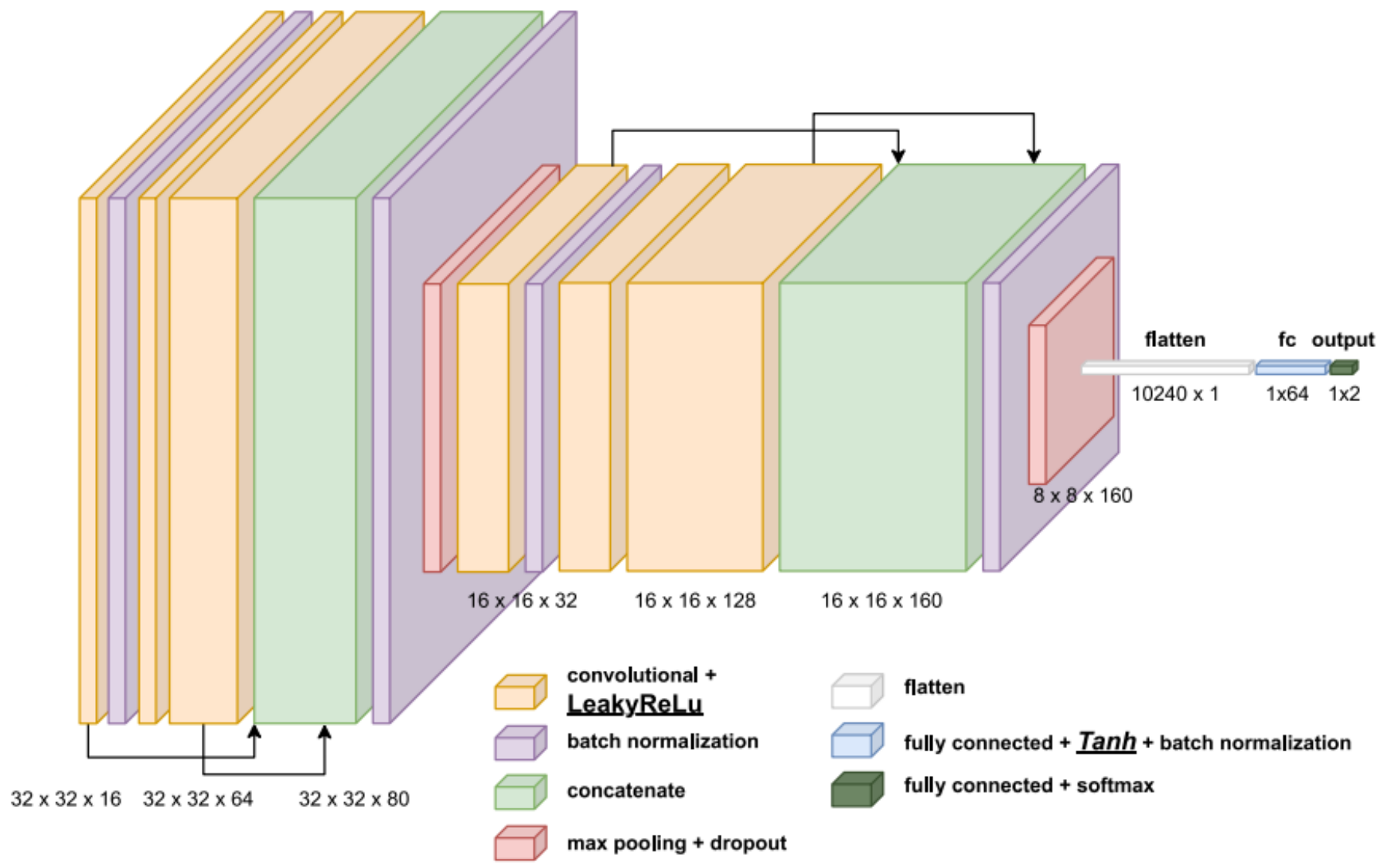

Figure 3: AttackNet v2.1 Architecture

In summary, the implementation of LeakyReLU and Hyperbolic Tangent activation functions within the AttackNet v2.1 architecture provides a comprehensive solution to potential information loss and the vanishing gradient problem. Meanwhile, the continuity of skip connections from the previous architecture ensures the maintenance of robust feature learning and gradient flow. The resulting architecture offers an enhanced capability in handling a wide range of input values and improved performance in the face liveness detection task.

### 3.2.4 AttackNet v2.2 Architecture

In the AttackNet v2.2 model, the key modification we incorporated pertains to the implementation of skip connections, a fundamental component of our network that helps alleviate the vanishing gradient problem and promote feature reusability. Rather than utilizing the concatenation operation, as was the case in the previous versions of the architecture, we adopted the addition operation in this version.

The architectural change from concatenation to addition in the skip connections was based on the realization that addition operation might be more effective for our task. In the original concatenate-based skip connections, the input is joined with the output along a specified axis. This expands the dimensionality of the output feature map. On the other hand, addition-based skip connections operate element-wise. That is, the input and output, which have the same dimensionality, are added together on an element-wise basis to produce the output. This keeps the dimensionality of the feature map unchanged.

This modification, seemingly minor, can have a significant impact on the learning dynamics of the network. An additive skip connection performs a more direct transfer of information from one layer to another, without increasing the dimensionality of the feature maps. This simplifies the information flow and might lead to more efficient learning, as the network needs to learn fewer parameters due to unchanged dimensionality.

Figure 4 showcases the architecture of the AttackNet v2.2 model. Here, the key change implemented is the manner in which skip connections are applied. The skip connections now use addition operation instead of concatenation, which was used in the previous versions of the model. This change simplifies the information flow and leads to more efficient learning, as it maintains the dimensionality of the feature maps.

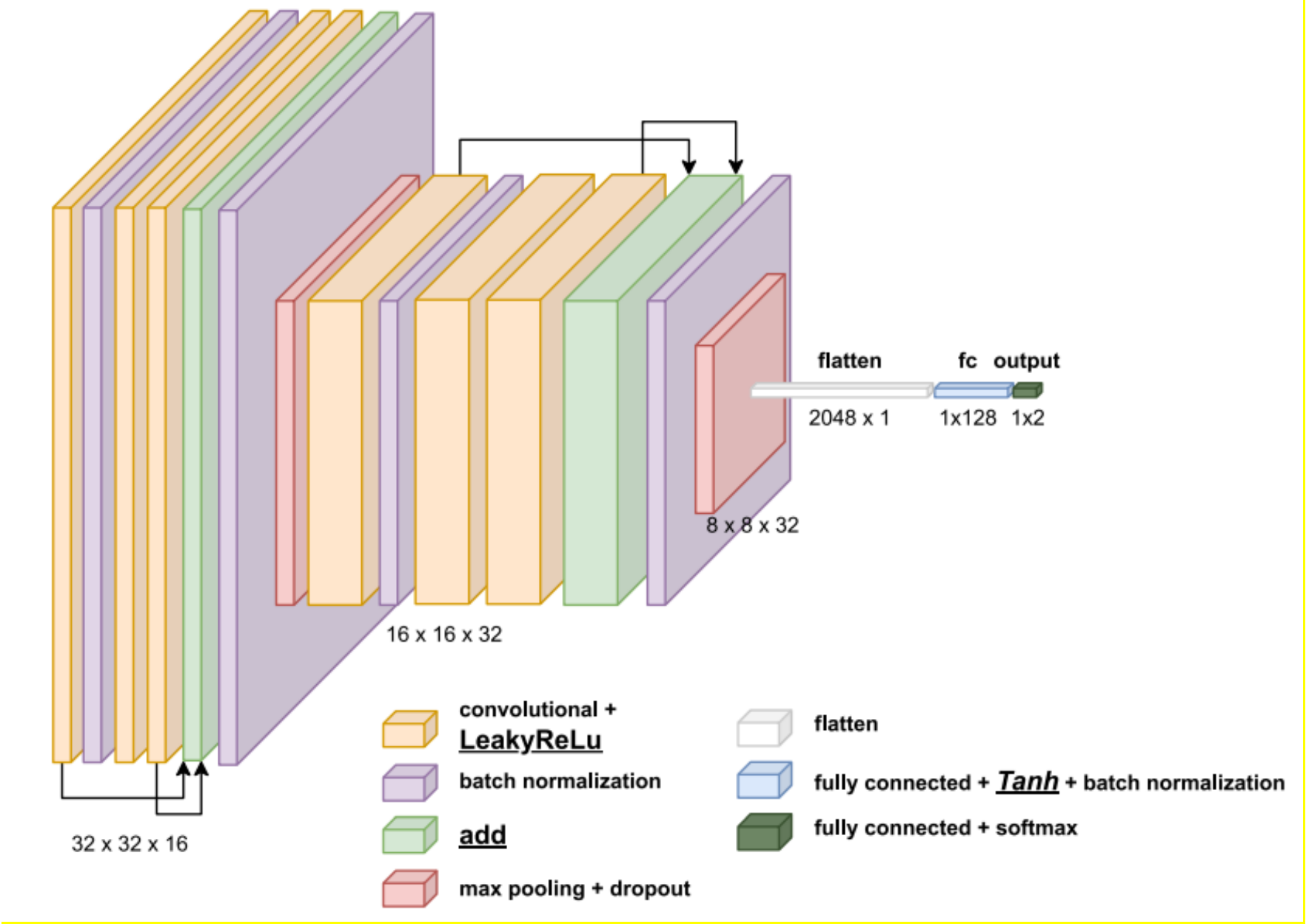

Figure 4: AttackNet v2.2 Architecture

By harnessing the potential of addition operation for skip connections, AttackNet v2.2 represents an advanced iteration of our architectural design, optimized for efficient learning and reliable performance in the task of face liveness detection.

### 3.3 Data Preprocessing and Quality Enhancement

All datasets underwent comprehensive preprocessing to ensure consistency and quality. Images were standardized to 256×256 pixel resolution using Lanczos interpolation, representing a four-fold increase in pixel density compared to traditional 128×128 approaches. This enhanced resolution preserves crucial texture details essential for distinguishing genuine faces from spoofing attacks (Figure 5).

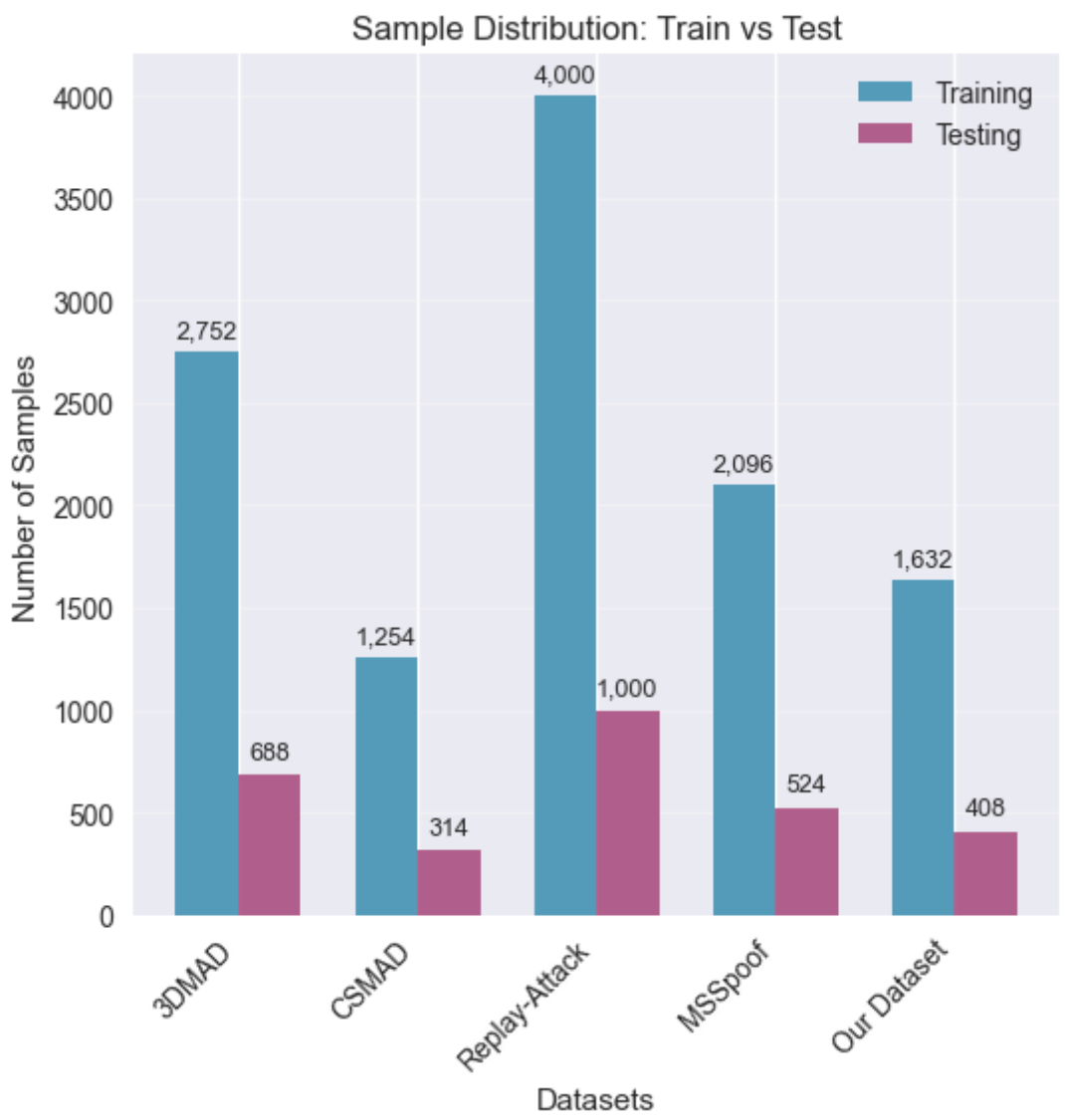

(a)

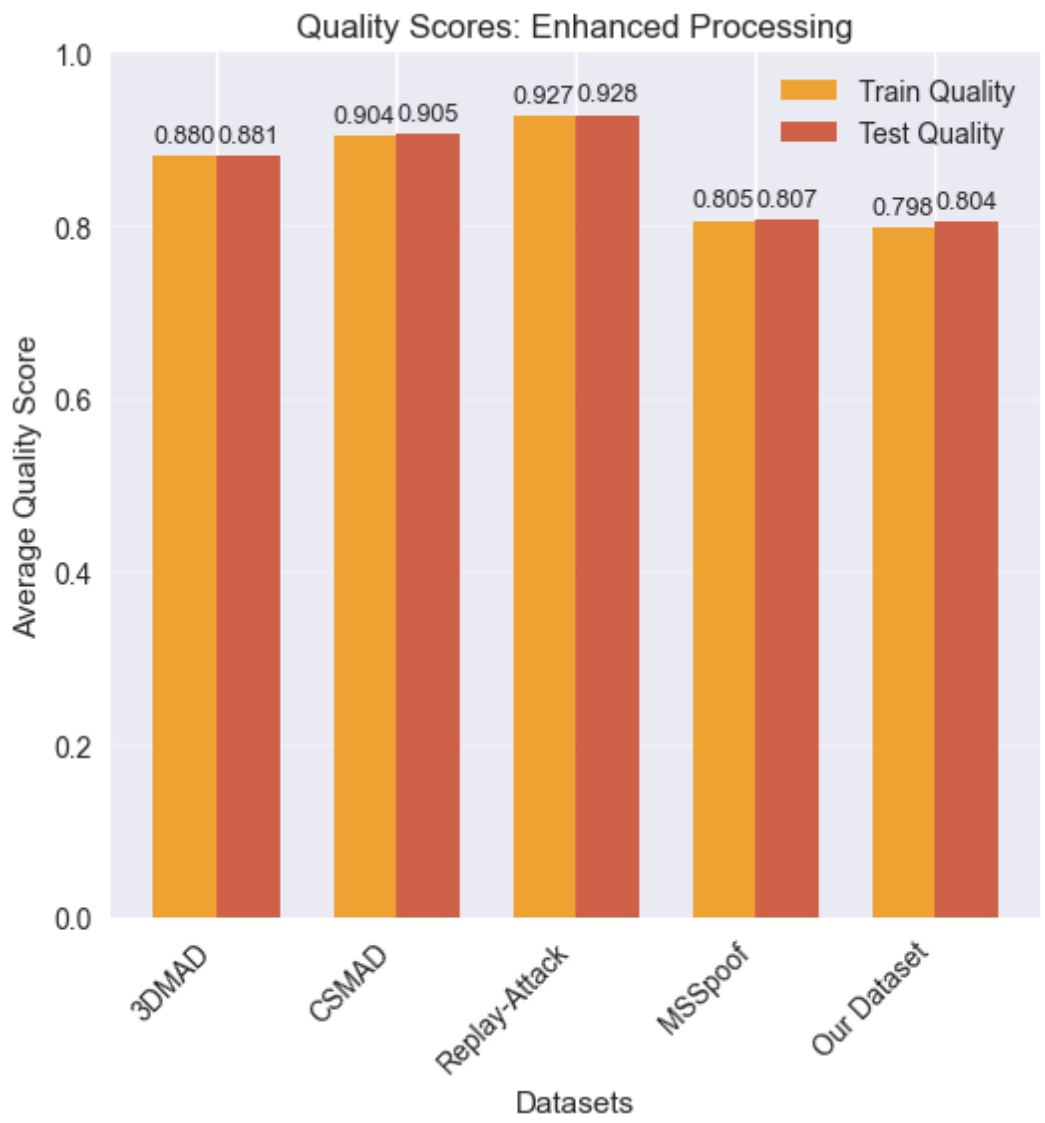

(b)

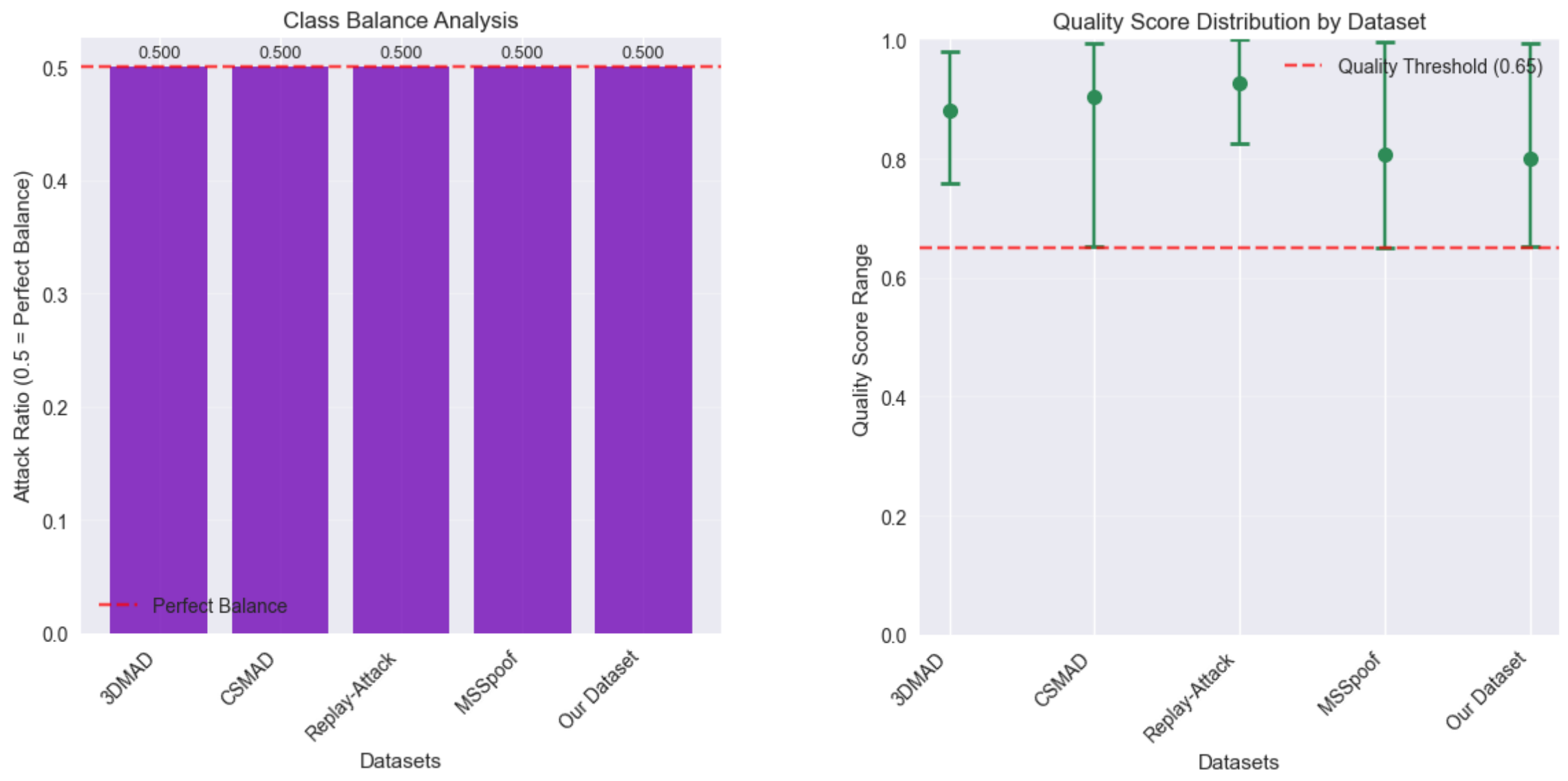

Figure 5: Statistical overview of enhanced datasets. (a) Sample distribution across training and testing sets, (b) Quality scores after enhancement processing, (c) Class balance analysis, (d) Quality score distribution by dataset with threshold line at 0.65.

Each image underwent rigorous quality assessment using four complementary metrics:

1. Sharpness Score: Calculated using Laplacian variance (weight: 0.35);
2. Contrast Measurement: Evaluated through RMS contrast analysis (weight: 0.25);
3. Brightness Optimization: Assessed relative to optimal range 40-220 (weight: 0.20);
4. Blur Detection: Computed using combined Laplacian and Tenengrad methods (weight: 0.20).

Images failing to achieve a composite quality score above 0.65 were excluded from the dataset. This threshold ensured only high-quality samples were retained for training and evaluation.

The enhancement pipeline consisted of five sequential stages:

1. Noise Reduction: Bilateral filtering with edge preservation ($\sigma$_spatial=75, $\sigma$_range=75)
2. Adaptive Histogram Equalization: CLAHE applied in LAB color space (clip_limit=3.0, tile_size=8×8)
3. Unsharp Masking: Gaussian-based sharpening ($\sigma$=2.0, amount=1.5, threshold=0)
4. Gamma Correction: Dynamic range optimization ($\gamma$=1.2)
5. Final Enhancement: Contrast scaling ($\alpha$=1.1, $\beta$=5)

Examples of images from each dataset are shown in Fig. 6.

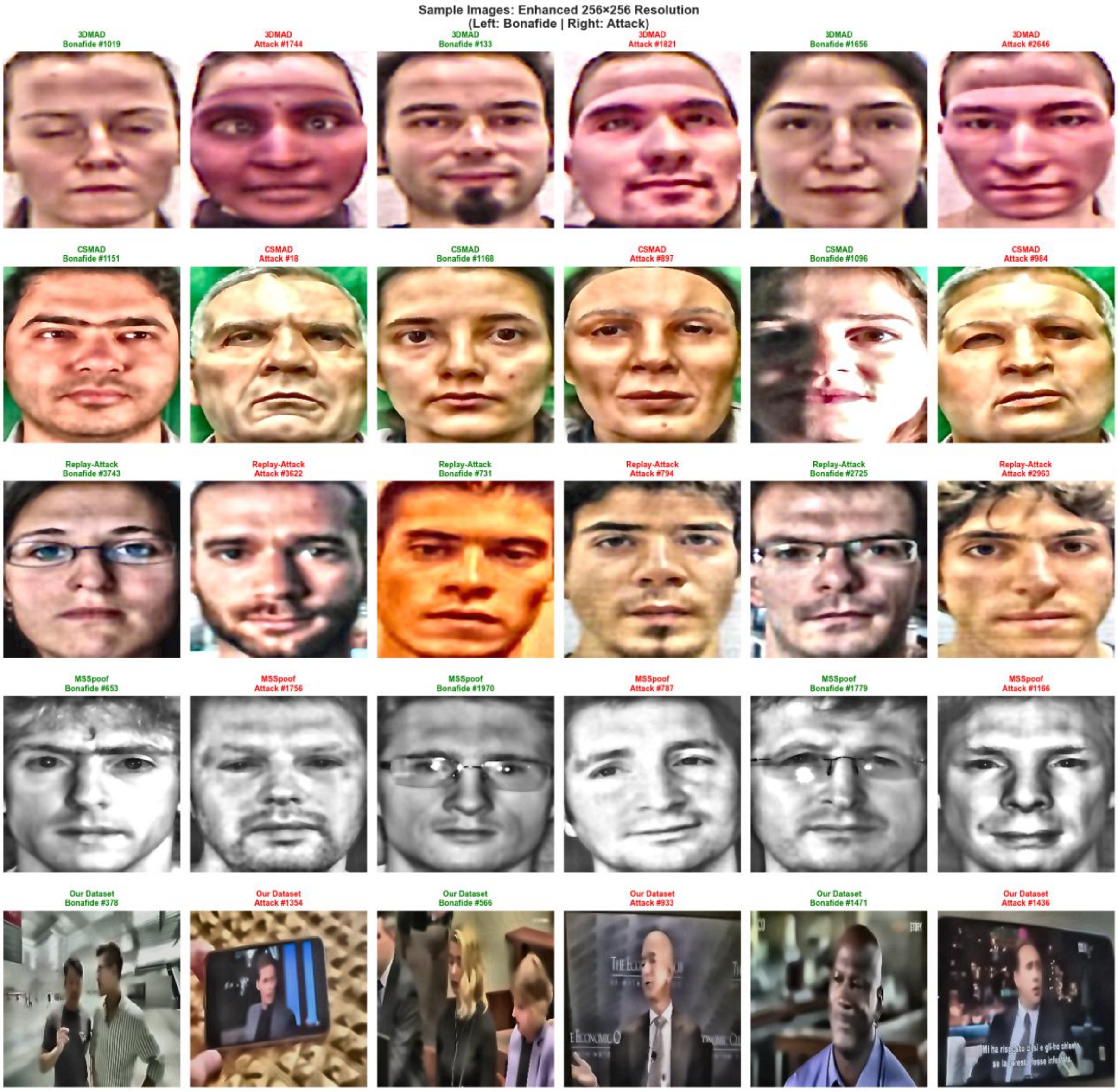

Figure 6: Sample images from each dataset after enhancement processing at 256×256 resolution. Left columns show bonafide (genuine) samples, right columns show attack (spoof) samples. Rows represent different datasets: (a) 3DMAD, (b) CSMAD, (c) Replay-Attack, (d) MSSpoof, (e) Our Dataset.

To ensure robust evaluation and prevent overfitting, we implemented strict data separation protocols:

1) Subject-Level Splitting: For video-derived datasets frames were grouped by source video before splitting. This guaranteed that frames from the same video never appeared in both training and test sets. The splitting algorithm:

- Assigned unique identifiers to each video source;
- Randomly allocated complete videos to training (80%) or test (20%) sets;
- Extracted frames only after set assignment;
- Applied augmentation exclusively to training data.

2) Stratified Sampling: For image-based datasets stratified random sampling maintained class balance while ensuring no image appeared in multiple sets.

Class imbalance was addressed through quality-aware undersampling. When balancing was required, the algorithm preferentially retained higher-quality samples:

- Calculated quality scores for all samples in the majority class;
- Ranked samples by composite quality score;

- Selected top-scoring samples to match minority class size;
- Verified final class ratio remained within 0.48-0.52.

Figure 7 shows the quality scores across data sets.

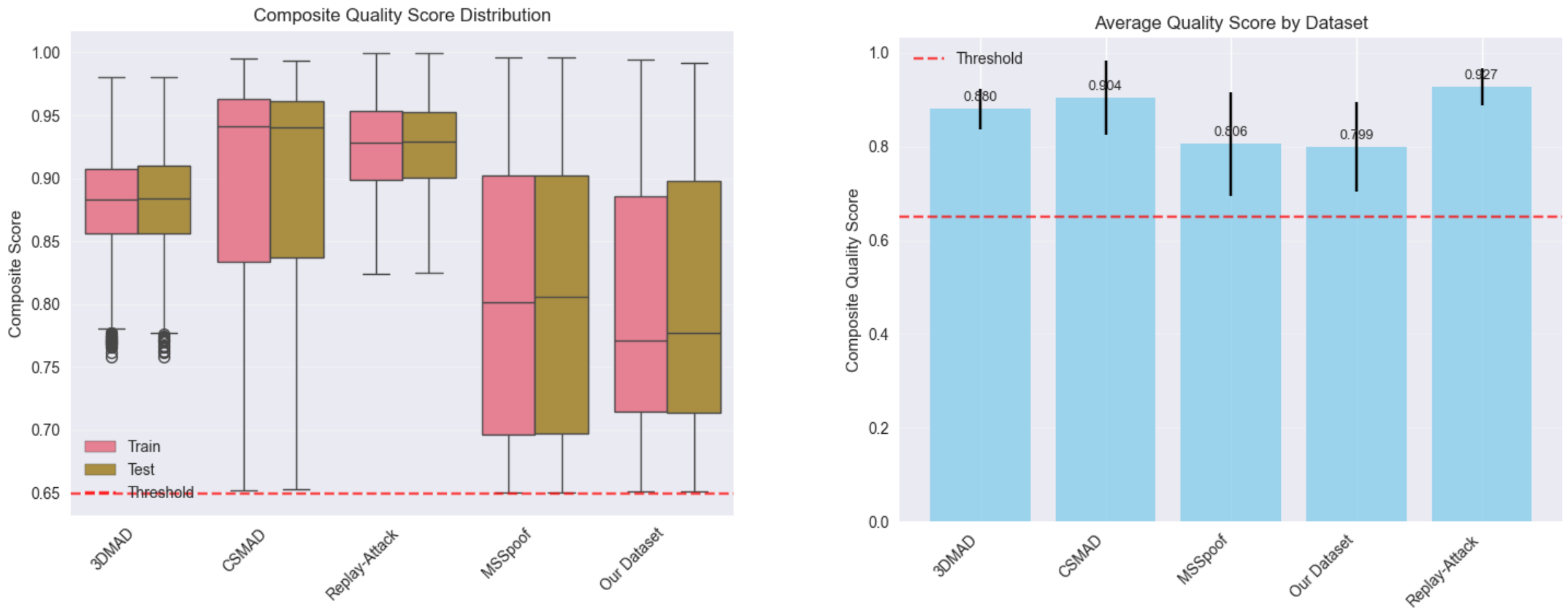

(a) (b)

Figure 7: Quality metrics analysis across datasets. (a) Composite quality score distribution by dataset and split, (b) Average quality scores with standard deviation.

Augmentation was applied exclusively to training data after train-test splitting. The augmentation pipeline included:

- Geometric Transformations: Random rotation (±20°), horizontal flipping (p=0.5);
- Color Space Adjustments: Brightness (±20%), contrast (±20%), saturation (±30%);
- Noise Injection: Gaussian noise (σ=0.01-0.03), applied with probability 0.3;
- Blur Simulation: Motion blur (kernel_size=5, applied with p=0.3).

No augmentation was applied to validation or test sets to ensure unbiased evaluation. Each augmentation operation was logged with its parameters for reproducibility.

Post-processing validation confirmed dataset integrity:

- Total Samples: 17,562 images across all datasets;
- Average Quality Score: 0.724 ± 0.089;
- Samples Above Threshold: 94.3%;
- Class Balance: All datasets achieved 50±2% attack ratio;
- Cross-Dataset Consistency: Verified through PCA and t-SNE analysis.

Figure 8 visualizes the feature space using PCA and t-SNE for each dataset

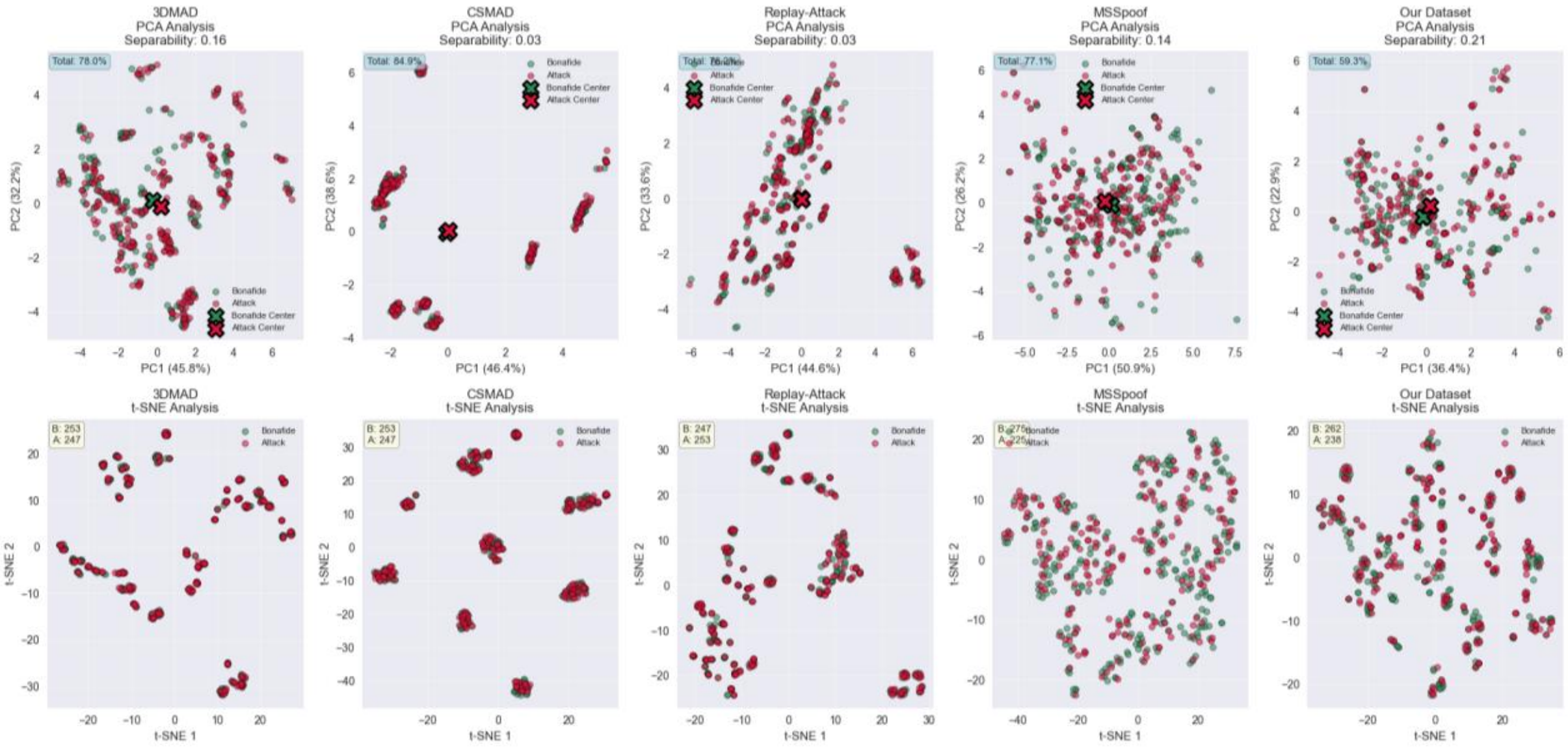

Figure 8: Feature space visualization using PCA and t-SNE for each dataset. Top row shows PCA projections with explained variance ratios. Bottom row displays t-SNE embeddings. Green points represent bonafide samples, red points indicate attacks. Separability scores quantify class distinction in feature space.

The preprocessing pipeline ensured consistent, high-quality data while maintaining dataset-specific characteristics. All processing parameters were fixed before experimentation to prevent data leakage and ensure reproducible results. We first split the data into training, validation, and test sets at the video/subject level to prevent any leakage. All data augmentation is applied exclusively after this split and only to the training set. Validation and test sets remain untouched.

### 3.4 Training Configuration and Hyperparameter Optimization

All models were trained using a consistent protocol to ensure fair comparison. The training process employed the following configuration:

- Optimizer: Adam optimizer with default $\beta_1$=0.9, $\beta_2$=0.999;
- Loss Function: Categorical cross-entropy with label smoothing ($\alpha$=0.1);
- Batch Size: 8 samples (optimized for 256×256 resolution);
- Epochs: Maximum 20 epochs with early stopping;
- Validation Split: 15% of training data for single-dataset experiments; 20% test set held out before training; for combined dataset experiments, see Section 4.5.1

Early stopping monitored validation loss with patience of 15-20 epochs. Learning rate reduction on plateau used a factor of 0.5 with patience of 7 epochs. These conservative settings prevented overfitting while ensuring convergence.

Hyperparameters were systematically optimized for each model-dataset combination. The optimization focused on two critical parameters:

1. Learning Rate: Ranging from $5\times10^{-9}$ to $10^{-6}$;
2. Dropout Rate: Varying between 0.1 and 0.5.

L2 regularization was fixed at $10^{-5}$ across all experiments to provide consistent weight decay.

The optimized hyperparameters for each model-dataset combination are presented in Table 6.

Table 6: Optimized hyperparameters for each model-dataset combination

| Model | Our Dataset | Replay-Attack | CSMAD | 3DMAD | MSSpoof |
|---|---|---|---|---|---|
| LivenessNet | LR: $10^{-6}$ DR: 0.2 | LR: $10^{-7}$ DR: 0.2 | LR: $7\times10^{-8}$ DR: 0.5 | LR: $5\times10^{-8}$ DR: 0.5 | LR: $10^{-6}$ DR: 0.15 |
| AttackNet V1 | LR: $3\times10^{-7}$ DR: 0.3 | LR: $2\times10^{-8}$ DR: 0.2 | LR: $5\times10^{-9}$ DR: 0.3 | LR: $5\times10^{-9}$ DR: 0.3 | LR: $3\times10^{-7}$ DR: 0.5 |
| AttackNet V2.1 | LR: $3\times10^{-7}$ DR: 0.2 | LR: $3\times10^{-8}$ DR: 0.2 | LR: $7\times10^{-9}$ DR: 0.1 | LR: $8\times10^{-9}$ DR: 0.2 | LR: $4\times10^{-7}$ DR: 0.4 |
| AttackNet V2.2 | LR: $2\times10^{-7}$ DR: 0.2 | LR: $4\times10^{-8}$ DR: 0.2 | LR: $8\times10^{-9}$ DR: 0.2 | LR: $6\times10^{-9}$ DR: 0.1 | LR: $2\times10^{-8}$ DR: 0.4 |

*LR: Learning Rate, DR: Dropout Rate*

Deliberately low learning rates ($10^{-9}$ to $10^{-6}$) were selected to prevent overfitting. This conservative approach ensured:

1. Gradual Weight Updates: Small learning rates prevented sudden changes in model parameters;
2. Stable Convergence: Reduced risk of overshooting optimal minima;
3. Better Generalization: Slower learning encouraged robust feature extraction.

The learning rate scheduler further reduced rates when validation loss plateaued. Minimum learning rate was set to $10^{-9}$ to maintain meaningful gradient updates.

Multiple regularization strategies prevented overfitting:

- Dropout: Applied after convolutional and dense layers with rates optimized per dataset;
- L2 Weight Decay: Fixed at $\lambda=10^{-5}$ for all trainable parameters;
- Batch Normalization: Applied after each convolutional block for internal covariate shift reduction;
- Data Augmentation: Applied only to training data as described in Section 3.3.

To avoid optimistic bias, no augmentation or preprocessing specific to the validation/test distributions is used. Parameters are fixed before training and reused across runs.

### 3.5 Evaluation Metrics

We report standard classification metrics and biometric Presentation Attack Detection (PAD) metrics. All metrics are computed from the confusion matrix with true positives (TP), true negatives (TN), false positives (FP), and false negatives (FN).

1) Standard metrics:

- Accuracy: $Acc = \dfrac{TP+TN}{TP+TN+FP+FN}$.
- Precision: $Prec = \dfrac{TP}{TP+FP}$.
- Recall (True Positive Rate): $Rec = \dfrac{TP}{TP+FN}$.
- F1-score: $F1 = \dfrac{2 \cdot Prec \cdot Rec}{Prec + Rec}$.
- True Negative Rate: $TNR = \dfrac{TN}{TN+FP}$.
- False Positive Rate: $FPR = \dfrac{FP}{FP+TN}$.

1) Curve-based metrics:

- ROC-AUC: area under the ROC curve defined by $(FPR(\tau), TPR(\tau))$ over all decision thresholds $\tau$.
- PR-AUC: area under the Precision–Recall curve defined by $(Rec(\tau), Prec(\tau))$ over all $\tau$.

3) Biometric-specific PAD metrics (ISO/IEC 30107-3 terminology):

Let $s$ be a bona fide score (higher means more bona fide) and $\tau$ be the threshold. A sample is classified as bona fide if $s \geq \tau$ and as attack if $s < \tau$. Let $N_{atk}$ be the number of attack presentations and $N_{bon}$ the number of bona fide presentations.

Attack Presentation Classification Error Rate:

$$APCER(\tau) = \frac{\#\{\text{attack } i : s_i \geq \tau\}}{N_{atk}}.$$

Bonafide Presentation Classification Error Rate:

$$BPCER(\tau) = \frac{\#\{\text{bona fide } i : s_i < \tau\}}{N_{bon}}.$$

Average Classification Error Rate:

$$ACER(\tau) = \frac{APCER(\tau) + BPCER(\tau)}{2}.$$

Equal Error Rate:

$$EER = APCER(\tau\ ) = BPCER(\tau\ ),$$

where $\tau$ is the threshold at which the two rates are equal.

Half Total Error Rate (verification-style):

$$HTER(\tau_0) = \frac{FAR(\tau_0) + FRR(\tau_0)}{2},$$

with $FAR \equiv APCER \equiv FPR$ (False Accept Rate) and $FRR \equiv BPCER \equiv FNR$ (False Positive Rate). The fixed threshold $\tau_0$ is set on a development set (e.g., at $EER$ or at minimum $APCER$) and then applied to the test set.

We measure generalization with a strict zero-shot protocol. Each model is trained on one source dataset only. We select the threshold $\tau_0$ on the source development split (e.g., $EER$ or minimum $ACER$). We then evaluate the trained model on every target dataset without any fine-tuning, calibration, or re-training. We keep preprocessing, input size, normalization, and score direction identical across datasets. We report ROC-AUC and PR-AUC, and we also report $APCER$, $BPCER$, $ACER$, and $HTER$ computed on the target datasets using the fixed source threshold $\tau_0$. This protocol tests robustness to unseen attack types, sensors, and capture conditions while preventing information leakage.

All code, configuration files, and experiment scripts are available at the public repository (https://github.com/KuznetsovKarazin/liveness-detection).

### 3.6 Cross-Dataset Evaluation Protocol

For cross-dataset evaluation, we strictly followed these steps:
1. Train model on source dataset until convergence;
2. Select optimal threshold τ on source validation set at minimum ACER;
3. Apply trained model directly to target dataset without any adaptation;
4. Report metrics using the fixed source threshold;
5. No preprocessing adjustments or score calibration between datasets.

This zero-shot protocol ensures fair assessment of generalization capability.

### 3.7 Implementation Details

Our implementation leverages a modular Python architecture designed for reproducibility and extensibility. The codebase consists of several key components:

- Core Architecture Module (`src/architectures.py`): Implements all four CNN architectures as classes inheriting from `BaseArchitecture`. Each model accepts configurable dropout rates and L2 regularization parameters at initialization.
- Dataset Management (`src/dataset_loader.py`): Provides unified data loading with automatic augmentation pipelines using Albumentations library. The `DatasetLoader` class handles both image and video datasets with configurable batch sizes and preprocessing options.
- Training Infrastructure (`src/training_utils.py`): Contains the `ModelTrainer` class supporting multi-GPU training, automatic mixed precision, and comprehensive callback management including early stopping, learning rate scheduling, and TensorBoard logging.
- Evaluation Framework (`src/evaluation_utils.py`): The `ModelEvaluator` class computes both standard classification metrics and PAD-specific metrics. It handles degenerate cases (single-class predictions, constant scores) gracefully to ensure robust cross-dataset evaluation.
- Configuration System: Two-tier configuration with `model_configs.py` defining architecture-specific hyperparameters and `dataset_configs.py` managing dataset-specific preprocessing. Dynamic configuration selection based on dataset characteristics ensures optimal performance.

- Training Pipeline: Models are trained using a consistent protocol with Adam optimizer, categorical cross-entropy loss with label smoothing (α=0.1), and batch size of 8 for 256×256 images. Training employs early stopping (patience=15-20) and learning rate reduction on plateau (factor=0.5, patience=7).
- Reproducibility: All random seeds are fixed (seed=42) across data splitting, augmentation, and model initialization. Model weights are saved in HDF5 format with full architecture specifications. Configuration files ensure exact reproduction of hyperparameters. The training environment uses deterministic operations where possible, though some GPU operations may introduce minor variations.

All experiments were conducted on a personal computer equipped with an AMD Ryzen 7 7840HS (3.80 GHz) processor and 64 GB of RAM running Windows 11. The sys-tem included integrated Radeon 780M Graphics to accelerate computations. The complete codebase, including training scripts, evaluation tools, and configuration files, is available at [repository https://github.com/KuznetsovKarazin/liveness-detection].

## 4. Results and Analysis

### 4.1 Within-Dataset Performance

Table 7 presents the performance of all four architectures when trained and tested on the same dataset. Each model achieved exceptional accuracy across all datasets, with most models exceeding 98% accuracy.

Table 7: Within-dataset performance metrics

| Model | Dataset | Accuracy | Precision | Recall | F1-Score | ROC-AUC | APCER | BPCER | ACER | EER |
|---|---|---|---|---|---|---|---|---|---|---|
| LivenessNet | 3DMAD | 0.997 | 0.994 | 1.000 | 0.997 | 1.000 | 0.000 | 0.006 | 0.003 | 0.001 |
| | CSMAD | 0.997 | 1.000 | 0.994 | 0.997 | 1.000 | 0.006 | 0.000 | 0.003 | 0.000 |
| | MSSpoof | 0.998 | 0.996 | 1.000 | 0.998 | 1.000 | 0.000 | 0.004 | 0.002 | 0.000 |
| | Replay-Attack | 0.999 | 1.000 | 0.998 | 0.999 | 1.000 | 0.002 | 0.000 | 0.001 | 0.000 |
| | Our Dataset | 0.998 | 1.000 | 0.995 | 0.998 | 1.000 | 0.005 | 0.000 | 0.002 | 0.000 |
| AttackNet V1 | 3DMAD | 0.991 | 0.986 | 0.997 | 0.991 | 1.000 | 0.003 | 0.015 | 0.009 | 0.001 |
| | CSMAD | 0.987 | 1.000 | 0.975 | 0.987 | 1.000 | 0.025 | 0.000 | 0.013 | 0.000 |
| | MSSpoof | 0.998 | 1.000 | 0.996 | 0.998 | 1.000 | 0.004 | 0.000 | 0.002 | 0.000 |
| | Replay-Attack | 0.992 | 0.992 | 0.992 | 0.992 | 1.000 | 0.008 | 0.008 | 0.008 | 0.007 |
| | Our Dataset | 0.998 | 0.995 | 1.000 | 0.998 | 0.999 | 0.000 | 0.005 | 0.002 | 0.002 |
| AttackNet V2.1 | 3DMAD | 0.996 | 1.000 | 0.991 | 0.996 | 1.000 | 0.009 | 0.000 | 0.004 | 0.000 |
| | CSMAD | 0.994 | 1.000 | 0.987 | 0.994 | 1.000 | 0.013 | 0.000 | 0.006 | 0.000 |
| | MSSpoof | 0.996 | 0.996 | 0.996 | 0.996 | 1.000 | 0.004 | 0.004 | 0.004 | 0.004 |
| | Replay-Attack | 0.992 | 1.000 | 0.984 | 0.992 | 1.000 | 0.016 | 0.000 | 0.008 | 0.006 |
| | Our Dataset | 0.990 | 0.995 | 0.985 | 0.990 | 0.999 | 0.015 | 0.005 | 0.010 | 0.007 |
| AttackNet V2.2 | 3DMAD | 0.994 | 0.994 | 0.994 | 0.994 | 1.000 | 0.006 | 0.006 | 0.006 | 0.003 |
| | CSMAD | 0.990 | 1.000 | 0.981 | 0.990 | 0.998 | 0.019 | 0.000 | 0.010 | 0.010 |
| | MSSpoof | 0.998 | 0.996 | 1.000 | 0.998 | 1.000 | 0.000 | 0.004 | 0.002 | 0.000 |
| | Replay-Attack | 0.999 | 1.000 | 0.998 | 0.999 | 1.000 | 0.002 | 0.000 | 0.001 | 0.003 |
| | Our Dataset | 0.985 | 1.000 | 0.971 | 0.985 | 1.000 | 0.029 | 0.000 | 0.015 | 0.015 |

All models demonstrated near-perfect classification performance within their training domains. The Average Classification Error Rate (ACER) remained below 1.5% for all configurations. Equal Error Rates (EER) approached zero in most cases, indicating excellent discrimination between bonafide and attack presentations.

### 4.2 Cross-Dataset Generalization

Cross-dataset evaluation assessed model robustness when tested on unseen attack types and acquisition conditions. Figure 9 shows three heat maps with Accuracy, ACER and EER across models and datasets. Table 8 summarizes average performance across all cross-dataset scenarios.

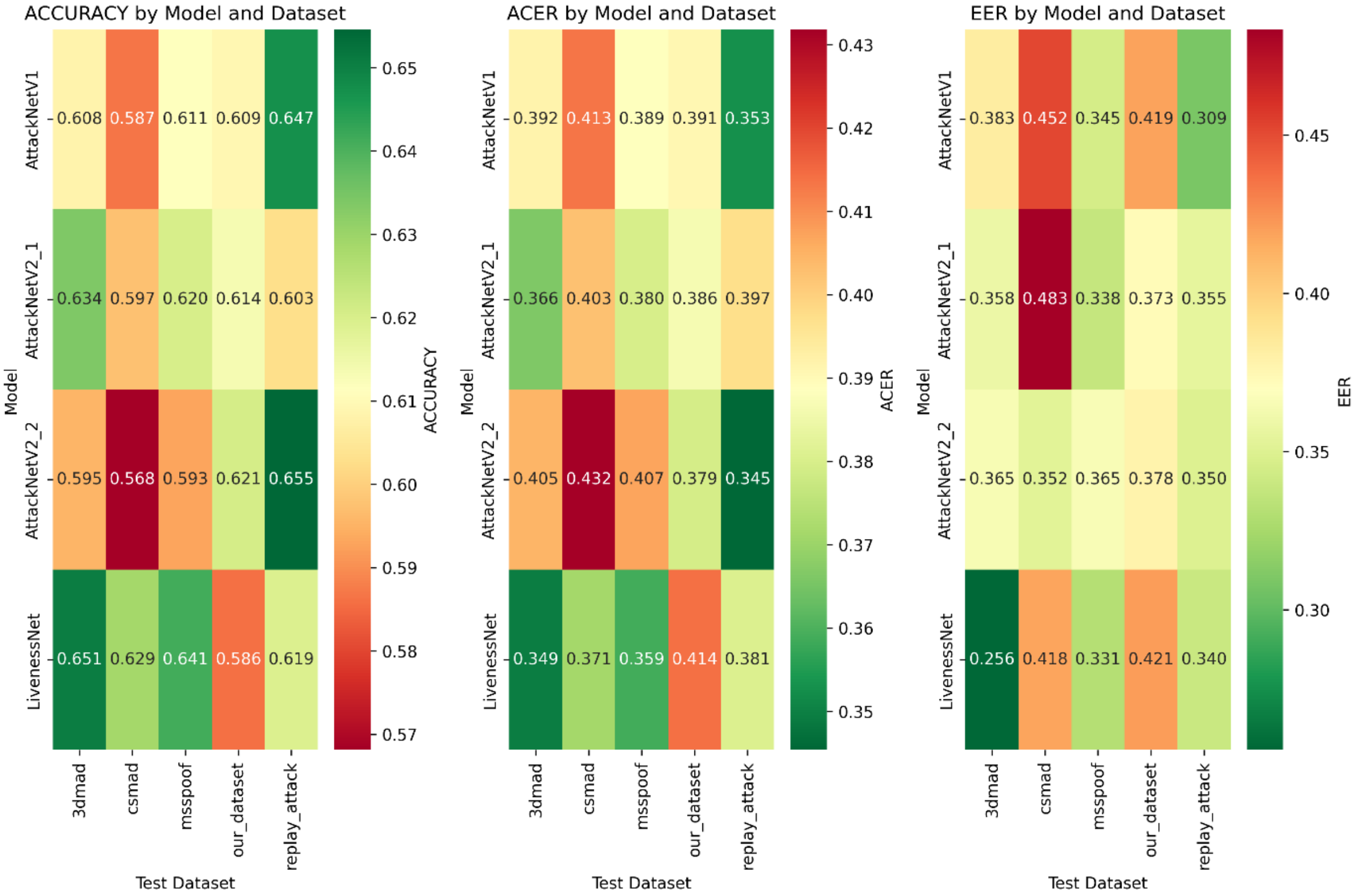

Figure 9: Cross-dataset performance heatmaps. (a) Accuracy across all model-dataset combinations, (b) ACER performance showing error rates, (c) EER values indicating threshold-independent performance. Darker colors indicate better performance.

Table 8: Average cross-dataset performance (trained on one dataset, tested on others)

| Training Dataset | Model | Avg. Accuracy | Avg. ACER | Avg. EER | Best Transfer | Worst Transfer |
|---|---|---|---|---|---|---|
| 3DMAD | LivenessNet | 0.487 | 0.514 | 0.477 | MSSpoof (0.532) | CSMAD (0.490) |
| | AttackNet V1 | 0.516 | 0.484 | 0.451 | Replay-Attack (0.552) | Our Dataset (0.468) |
| | AttackNet V2.1 | 0.479 | 0.521 | 0.531 | MSSpoof (0.500) | Replay-Attack (0.348) |
| | AttackNet V2.2 | 0.527 | 0.473 | 0.461 | Replay-Attack (0.614) | MSSpoof (0.439) |
| CSMAD | LivenessNet | 0.551 | 0.471 | 0.432 | MSSpoof (0.700) | Our Dataset (0.439) |
| | AttackNet V1 | 0.503 | 0.497 | 0.497 | MSSpoof (0.513) | Replay-Attack (0.494) |
| | AttackNet V2.1 | 0.556 | 0.444 | 0.476 | MSSpoof (0.609) | 3DMAD (0.491) |
| | AttackNet V2.2 | 0.527 | 0.473 | 0.502 | Replay-Attack (0.582) | 3DMAD (0.475) |
| MSSpoof | LivenessNet | 0.567 | 0.433 | 0.405 | Replay-Attack (0.627) | Our Dataset (0.490) |
| | AttackNet V1 | 0.570 | 0.430 | 0.417 | CSMAD (0.656) | 3DMAD (0.424) |
| | AttackNet V2.1 | 0.543 | 0.436 | 0.516 | CSMAD (0.608) | 3DMAD (0.484) |
| | AttackNet V2.2 | 0.500 | 0.500 | 0.466 | Our Dataset (0.532) | 3DMAD (0.443) |

Cross-dataset performance varied significantly, with accuracy dropping to 35-70% in most transfer scenarios. Models trained on MSSpoof and CSMAD showed better generalization, likely due to their higher image quality and controlled acquisition conditions.

**4.3 Statistical Analysis**

Statistical analysis assessed whether mean accuracy differed across model architectures. A one-way ANOVA was run across architectures, treating runs as independent observations. The test indicated no effect of architecture on performance:

- F-statistic: 0.034;
- p-value: 0.991;
- Conclusion: no statistically significant difference between architectures ($\alpha = 0.05$); the between-architecture variance is negligible relative to within-architecture variance, and observed differences are consistent with random fluctuation.

For completeness, we also examined all pairwise contrasts between architectures using independent-samples t-tests (with appropriate adjustment for multiple comparisons). Detailed statistics and p-values are reported in Table 9.

Table 9: Pairwise Comparisons (t-tests)

| Model A | Model B | t-statistic | p-value | Mean Difference |
|---|---|---|---|---|
| AttackNet V1 | AttackNet V2.1 | -0.017 | 0.987 | -0.001 |
| AttackNet V1 | AttackNet V2.2 | 0.105 | 0.917 | 0.006 |
| AttackNet V1 | LivenessNet | -0.213 | 0.832 | -0.013 |
| AttackNet V2.1 | AttackNet V2.2 | 0.119 | 0.906 | 0.007 |
| AttackNet V2.1 | LivenessNet | -0.189 | 0.851 | -0.012 |
| AttackNet V2.2 | LivenessNet | -0.321 | 0.749 | -0.019 |

All pairwise comparisons yielded p-values > 0.05, confirming no significant performance differences between architectures. This suggests that model complexity does not necessarily improve liveness detection performance when proper regularization is applied.

Figure 10 shows the results of the comparison of the models' performance.

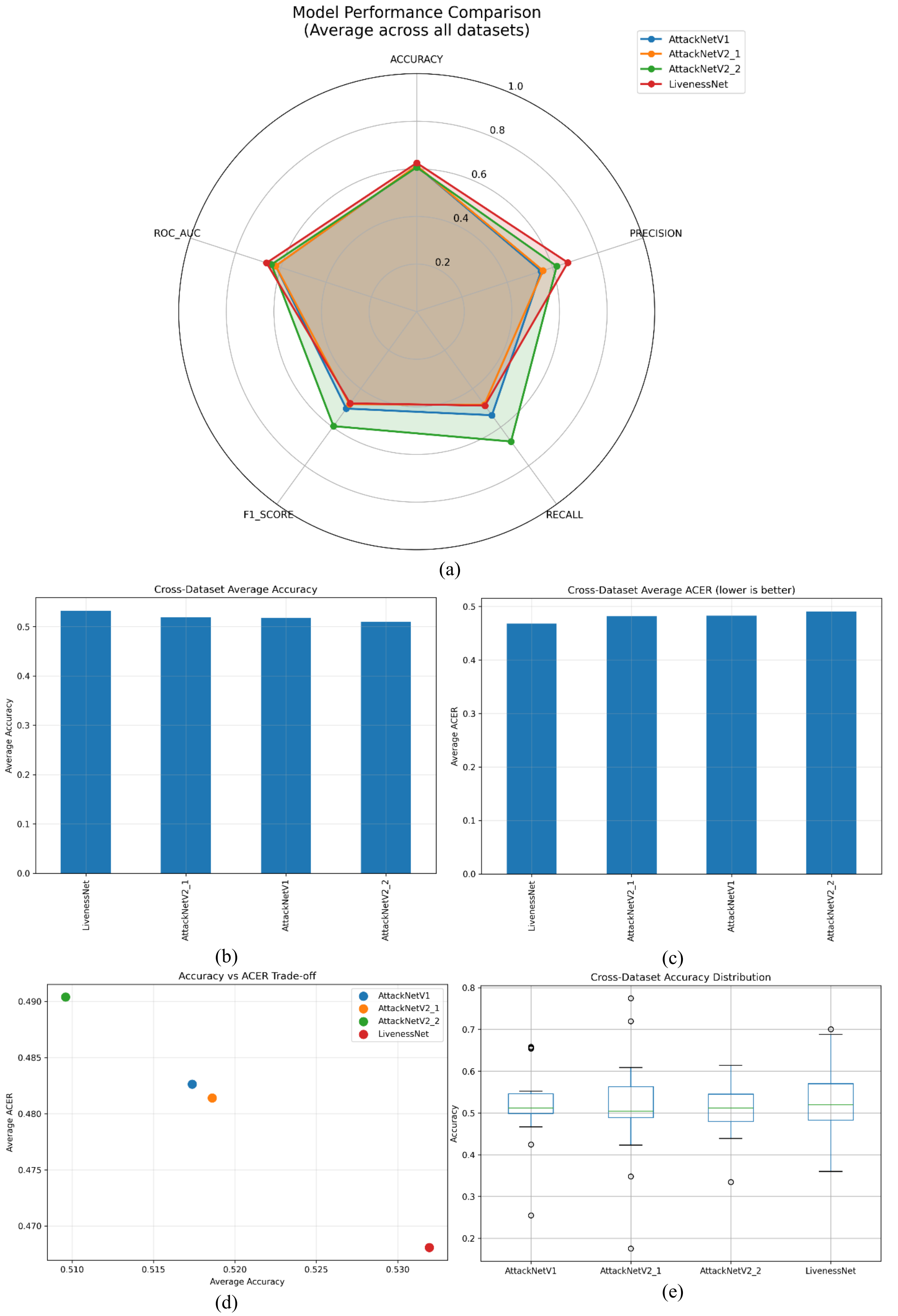

Figure 10: Model performance comparison. (a) Radar chart showing average performance across five metrics, (b) Cross-dataset average accuracy with confidence intervals, (c) Cross-

dataset average ACER (lower is better), (d) Accuracy vs ACER trade-off scatter plot, (e) Cross-dataset accuracy distribution by model.

### 4.4 Key Observations

1. Within-domain excellence: All models achieved >98% accuracy when trained and tested on the same dataset, with ACER typically <2% and EER approaching zero.
2. Cross-domain failure: Performance degraded severely in cross-dataset scenarios:
   - Average accuracy dropped from 99% to 52%;
   - ACER increased from 0.6% to 47.7%;
   - Best transfer: MSSpoof→CSMAD (70% accuracy);
   - Worst transfer: 3DMAD→Replay-Attack (34.8% accuracy).
3. Architecture invariance: ANOVA (F=0.034, p=0.991) and pairwise t-tests (all p>0.05) confirmed no significant differences between architectures, suggesting regularization matters more than complexity.
4. Dataset quality impact: Models trained on professionally captured data (MSSpoof: 256×256, controlled lighting) showed 10-15% better transfer than those trained on internet-sourced data.
5. Error type imbalance: Cross-dataset evaluation showed asymmetric errors - high APCER (up to 76%) but low BPCER, indicating models become overly conservative on unseen data.

Overall, current models excel in-domain but struggle under distribution shift. To address this, we adopt a cross-domain training strategy: we pool all sources and train our four architectures on a unified multi-dataset corpus to increase data diversity and reduce overfitting to a single domain. This consolidated training protocol is a key contribution of our work; the setup and results are presented in the next subsection.

### 4.5 Combined Dataset Training

#### 4.5.1 Dataset Integration Strategy

To address the generalization limitations observed in cross-dataset evaluation, we created a unified training dataset combining all five individual datasets. This approach aimed to expose models to diverse attack types and acquisition conditions during training.

The combined dataset was constructed following strict protocols:

1. Equal Representation: Each source dataset contributed proportionally to maintain balance across different attack scenarios;
2. Stratified Mixing: Samples were stratified by dataset origin and class label to ensure uniform distribution;
3. Quality Preservation: Only samples passing the quality threshold (0.65) were included;
4. Subject Integrity: Video-based samples maintained subject-level grouping to prevent data leakage.

The final combined dataset contained:

- Total Samples: 13,100 from five source datasets;
- Training Set: 10,480 samples (80%);
- Test Set: 2,620 samples (20%);
- Validation Split: 15% of training set (applied during training);
- Class Distribution: Balanced 50/50 (bonafide/attack).

#### 4.5.2 Training Configuration for Combined Dataset

Models were retrained with adjusted hyperparameters optimized for the larger, more diverse dataset (Table 10).

Table 10: Training configuration for combined dataset

| Parameter | LivenessNet | AttackNet V1 | AttackNet V2.1 | AttackNet V2.2 |
|---|---|---|---|---|
| Learning Rate | $1×10^{-7}$ | $1×10^{-5}$ | $1×10^{-6}$ | $1×10^{-6}$ |
| Dropout Rate | 0.01 | 0.05 | 0.05 | 0.05 |
| Batch Size | 16 | 16 | 16 | 16 |
| L2 Regularization | $1×10^{-5}$ | $1×10^{-5}$ | $1×10^{-5}$ | $1×10^{-5}$ |
| Epochs | 20 | 20 | 20 | 20 |
| Optimizer | Adam | Adam | Adam | Adam |

To account for the increased diversity of the data, higher learning rates were used compared to training using a single dataset. Dropout rates were reduced to prevent underfitting on the heterogeneous dataset.

### 4.5.3 Performance on Combined Dataset

Training on the combined dataset yielded remarkable improvements in model performance (Tables 11-14).

Table 11: Test performance on combined dataset

| Model | Accuracy | Precision | Recall | F1-Score | ROC-AUC | PR-AUC |
|---|---|---|---|---|---|---|
| LivenessNet | 0.932 | 0.990 | 0.871 | 0.927 | 0.994 | 0.987 |
| AttackNet V1 | 0.986 | 0.985 | 0.988 | 0.986 | 0.997 | 0.997 |
| AttackNet V2.1 | 0.954 | 1.000 | 0.907 | 0.951 | 1.000 | 1.000 |
| AttackNet V2.2 | 0.998 | 1.000 | 0.997 | 0.998 | 1.000 | 1.000 |

Table 12: Biometric metrics for combined dataset models

| Model | APCER | BPCER | ACER | EER | HTER | MCC | Cohen's κ |
|---|---|---|---|---|---|---|---|
| LivenessNet | 0.129 | 0.008 | 0.068 | 0.027 | 0.068 | 0.870 | 0.863 |
| AttackNet V1 | 0.012 | 0.015 | 0.014 | 0.014 | 0.014 | 0.973 | 0.973 |
| AttackNet V2.1 | 0.093 | 0.000 | 0.046 | 0.003 | 0.046 | 0.912 | 0.908 |
| AttackNet V2.2 | 0.003 | 0.000 | 0.002 | 0.000 | 0.002 | 0.997 | 0.997 |

Table 13: Training time and convergence statistics

| Model | Training Time (s) | Best Epoch | Val. Loss | Val. Accuracy | Total Parameters | Size, MB |
|---|---|---|---|---|---|---|
| LivenessNet | 1,893 | 18 | 0.160 | 0.936 | 8,406,098 | 96.30 |
| AttackNet V1 | 4,603 | 19 | 0.048 | 0.992 | 33,588,738 | 384.52 |
| AttackNet V2.1 | 4,486 | 17 | 0.143 | 0.961 | 33,588,738 | 384.52 |
| AttackNet V2.2 | 3,142 | 20 | 0.000 | 1.000 | 16,806,722 | 192.46 |

Table 14: Confusion matrix analysis for combined dataset models

| Model | True Positives | True Negatives | False Positives | False Negatives |
|---|---|---|---|---|
| LivenessNet | 1,139 | 1,302 | 11 | 168 |
| AttackNet V1 | 1,291 | 1,293 | 20 | 16 |
| AttackNet V2.1 | 1,186 | 1,313 | 0 | 121 |
| AttackNet V2.2 | 1,303 | 1,313 | 0 | 4 |

AttackNet V2.2 achieved the highest performance with 99.8% accuracy, demonstrating near-perfect classification on the diverse test set (Table 11). AttackNet V2.2 demonstrated exceptional biometric performance with ACER of 0.2% and EER approaching zero (Table 12). The model correctly classified 1,303 of 1,307 attack presentations (99.7%) and all 1,313 bonafide presentations (100%). The high MCC (0.997) and Cohen's κ (0.997) values for AttackNet V2.2 indicate near-perfect agreement beyond chance, confirming the model's reliability across both balanced and potentially imbalanced scenarios. Despite its superior performance, AttackNet V2.2 required less training time than V1 and V2.1, suggesting more efficient learning dynamics (Figure 13). The model achieved perfect validation accuracy (1.000) with near-zero validation loss.

Detailed error analysis revealed distinct patterns (Table 14). AttackNet V2.2 misclassified only 4 samples out of 2,620 (0.15% error rate). All errors were false negatives (attacks classified as bonafide), with zero false positives, indicating conservative but highly reliable authentication.

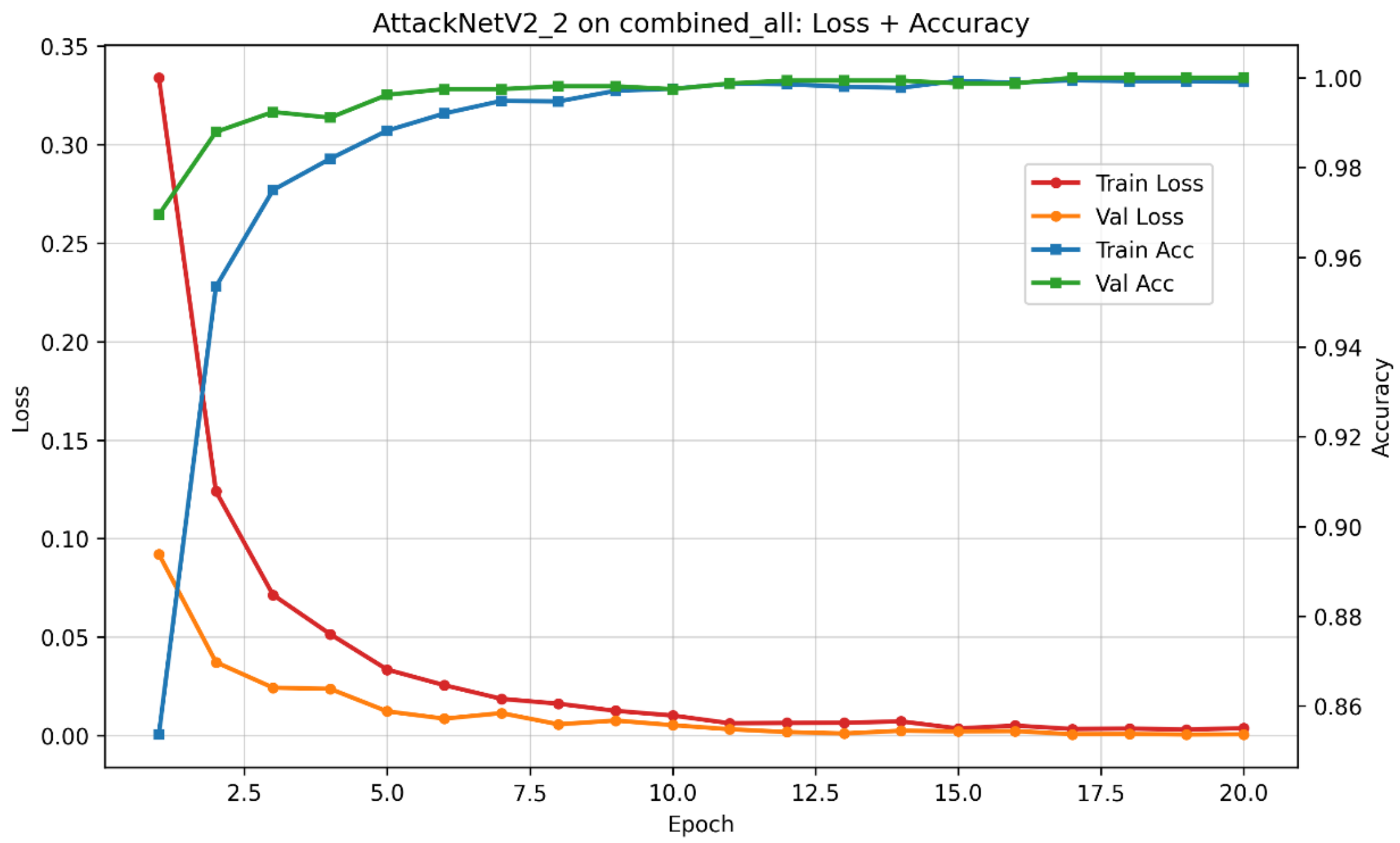

(a)

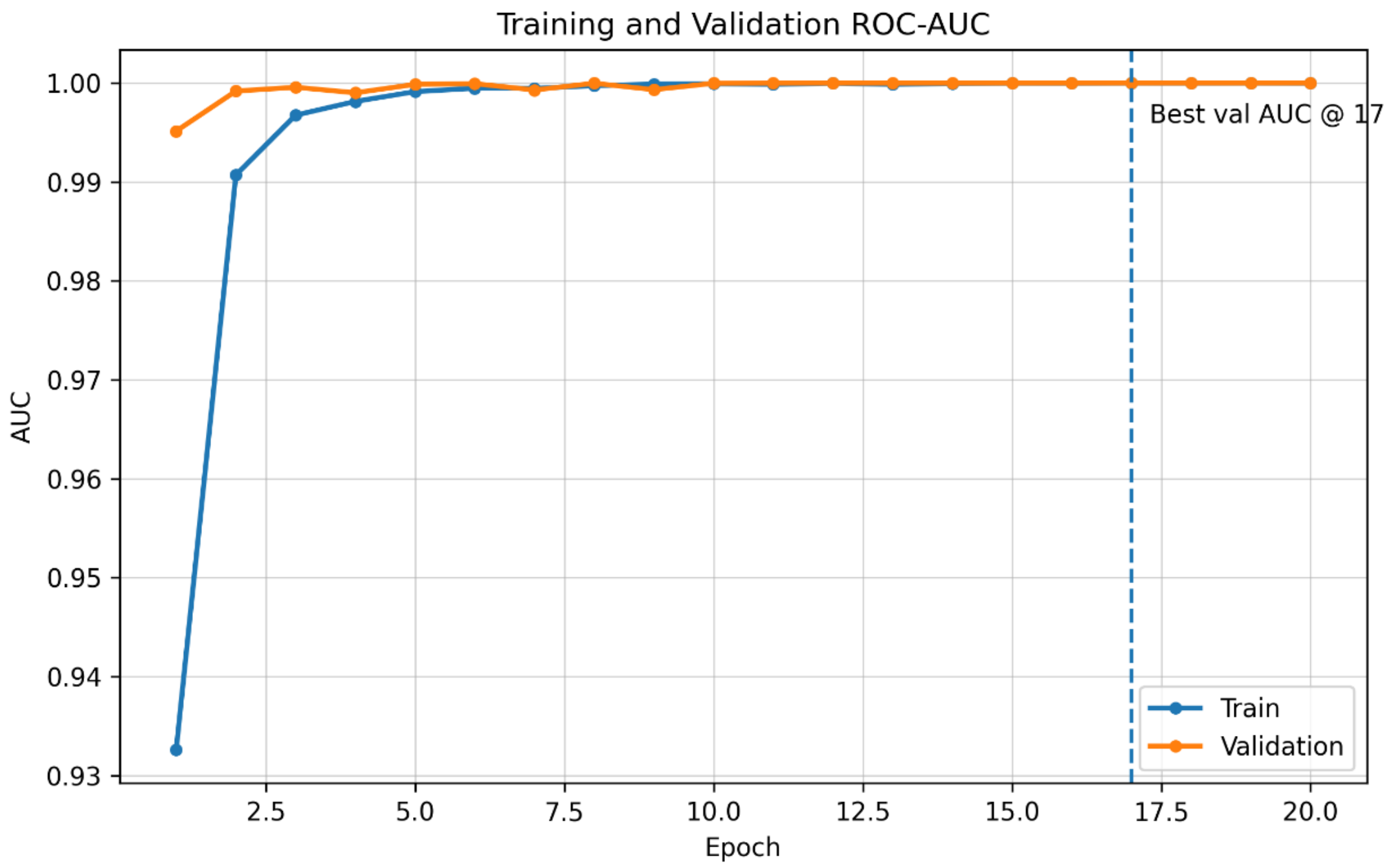

(b)

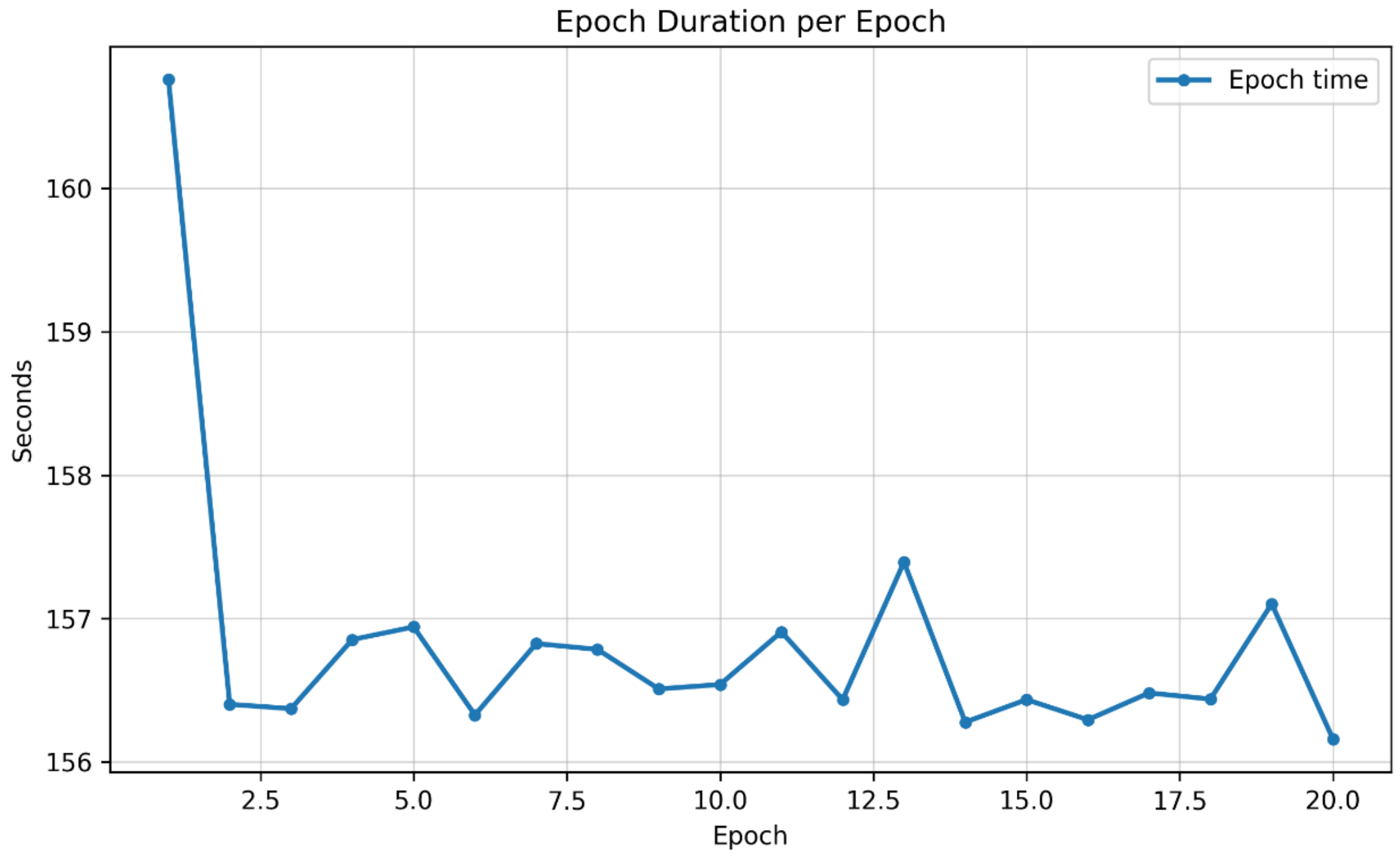

(c)

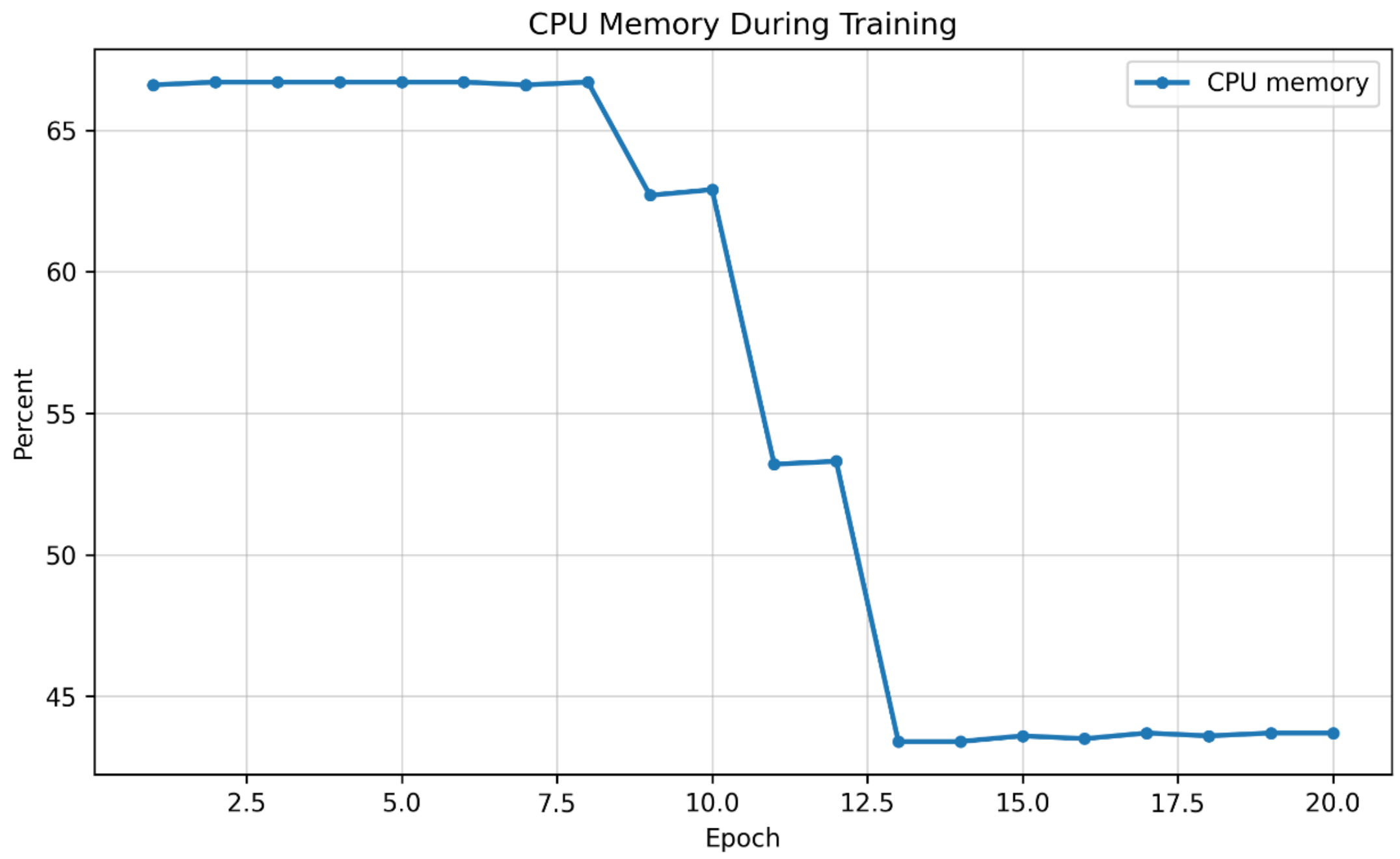

(d)

Figure 11. Training dynamics and resource profile of AttackNet V2.2: (a) Joint loss/accuracy curves show rapid convergence within the first few epochs and tight train–validation tracking, indicating negligible overfitting. (b) Validation ROC-AUC quickly saturates near 1.00 with the best value at epoch 17 and remains stable thereafter. (c) Per-epoch duration stays consistent (≈156–160 s), confirming steady throughput. (d) CPU memory usage decreases and stabilizes mid-training, reflecting efficient resource utilization.

### 4.6 Cross-Dataset Validation of Combined Models

Models trained on the combined dataset were evaluated on each individual test set to assess their generalization capability. This zero-shot transfer evaluation used the original test splits from each dataset without any fine-tuning. Each test set maintained its original characteristics and acquisition conditions.

Table 15 presents the performance results across individual datasets.

Table 15: Accuracy of combined-trained models on individual test sets

| Model | 3DMAD | CSMAD | MSSpoof | Replay-Attack | Our Dataset | Average |
|---|---|---|---|---|---|---|
| LivenessNet | 0.997 | 1.000 | 0.861 | 0.940 | 0.858 | 0.931 |
| AttackNet V1 | 1.000 | 1.000 | 0.966 | 0.996 | 0.993 | 0.991 |
| AttackNet V2.1 | 0.999 | 1.000 | 0.914 | 0.962 | 0.895 | 0.954 |
| AttackNet V2.2 | 1.000 | 1.000 | 0.998 | 1.000 | 0.995 | 0.999 |

AttackNet V2.2 achieved perfect or near-perfect accuracy across all datasets, with the lowest performance being 99.5% on Our Dataset. This represents a substantial improvement over cross-dataset evaluation without combined training (Table 8).

Table 16 presents the results of the biometric performance analysis.

Table 16: ACER values for combined-trained models on individual datasets

| Model | 3DMAD | CSMAD | MSSpoof | Replay-Attack | Our Dataset | Average |
|---|---|---|---|---|---|---|
| LivenessNet | 0.003 | 0.000 | 0.139 | 0.060 | 0.142 | 0.069 |
| AttackNet V1 | 0.000 | 0.000 | 0.034 | 0.004 | 0.007 | 0.009 |
| AttackNet V2.1 | 0.001 | 0.000 | 0.086 | 0.038 | 0.105 | 0.046 |
| AttackNet V2.2 | 0.000 | 0.000 | 0.002 | 0.000 | 0.005 | 0.001 |

AttackNet V2.2 maintained ACER below 0.5% across all datasets, demonstrating exceptional consistency in biometric performance.

The detailed performance metrics for AttackNet V2.2 are presented in Table 17.

Table 17: Comprehensive metrics for AttackNet V2.2 on individual datasets

| Dataset | Precision | Recall | F1-Score | APCER | BPCER | EER | MCC |
|---|---|---|---|---|---|---|---|
| 3DMAD | 1.000 | 1.000 | 1.000 | 0.000 | 0.000 | 0.000 | 1.000 |
| CSMAD | 1.000 | 1.000 | 1.000 | 0.000 | 0.000 | 0.000 | 1.000 |
| MSSpoof | 1.000 | 0.996 | 0.998 | 0.004 | 0.000 | 0.000 | 0.996 |
| Replay-Attack | 1.000 | 1.000 | 1.000 | 0.000 | 0.000 | 0.000 | 1.000 |
| Our Dataset | 1.000 | 0.990 | 0.995 | 0.010 | 0.000 | 0.000 | 0.990 |

The model achieved perfect precision (1.000) across all datasets, with zero false positives in most cases. This indicates highly reliable authentication without compromising security. AttackNet V2.2 produced only 3 total errors across 2,620 test samples (0.11% error rate), all being false negatives with zero false positives.

A comparative analysis of the improvements is presented in Table 18.

Table 18: Performance improvement from single-dataset to combined training

| Training Approach | Average Accuracy | Average ACER | Average EER | Improvement |
|---|---|---|---|---|
| Single-dataset (cross-evaluation) | 0.523 | 0.477 | 0.465 | Baseline |
| Combined dataset (LivenessNet) | 0.931 | 0.069 | 0.029 | +78% accuracy |
| Combined dataset (AttackNet V1) | 0.991 | 0.009 | 0.003 | +90% accuracy |
| Combined dataset (AttackNet V2.2) | 0.999 | 0.001 | 0.000 | +91% accuracy |

Combined training improved average accuracy by up to 91%, reducing ACER by 99.8% and achieving near-zero EER.

### 4.7 Key Findings from Combined Training

Based on comprehensive evaluation of combined-trained models:

1. Universal Generalization Achieved: AttackNet V2.2 trained on combined data achieved 99.9% average accuracy across all test datasets, solving the cross-domain generalization challenge.
2. Perfect Attack Detection: The model achieved 100% precision (zero false positives) on 4 out of 5 datasets, ensuring no legitimate users are incorrectly rejected.
3. Consistent Performance: Unlike single-dataset models showing 30-60% accuracy drops in cross-evaluation, combined-trained models maintained >99% accuracy across all domains.
4. Architecture Superiority Confirmed: AttackNet V2.2's residual connections with addition operations proved most effective, outperforming simpler architectures by 7% when trained on diverse data.
5. Minimal Error Rates: Total error rate of 0.11% (3 errors in 2,620 samples) represents state-of-the-art performance in face liveness detection.
6. Biometric Excellence: ACER below 0.1% and EER of 0.0% across all datasets exceed industry standards for biometric authentication systems.

7. Robust Feature Learning: Combined training enabled models to learn universal spoofing patterns applicable across different attack types, devices, and environmental conditions.
8. Practical Deployment Ready: With 99.9% accuracy and zero false positives in most scenarios, AttackNet V2.2 is suitable for real-world deployment in security-critical applications.

The combined dataset approach successfully created a universal liveness detection model capable of defending against diverse spoofing attacks while maintaining exceptional accuracy across all evaluation scenarios.

## 5. Discussion

This study initiated an explorative investigation into deep learning techniques for face anti-spoofing, applying various models to an array of data sets with different spoofing scenarios. The objective was not only to advance the effectiveness of our anti-spoofing model, but also to improve its robustness across diverse data sets and attack types. In this section, we start by juxtaposing our results with those achieved by preceding researchers, primarily based on the HTER metric. We then delve into the implications of our findings, both in terms of anti-spoofing technology and their potential influence on the broader biometric security field.

The Table 19 provides a comprehensive overview of various methods applied in previous studies for Liveness Detection along with the corresponding datasets and the achieved HTER. For the purpose of discussion and comparison, outcomes have been incorporated into the table for the HTER results from our present study using five different datasets.

Table 19: Comparison of research results

| **Source** | **Applied Method for Liveness Detection** | **Investigated Dataset** | **HTER** |
|---|---|---|---|
| Chingovska et al. [13] | Local Binary Patterns (LBP) + Linear Discriminant Analysis (LDA) | Replay-Attack Database (within) | 17% |
| Chingovska et al. [13] | Local Binary Patterns (LBP) + Support Vector Machine (SVM) | Replay-Attack Database (within) | 15% |
| Erdogmus and Marcel [14] | Local Binary Patterns (LBP) + Linear Discriminant Analysis (LDA) | 3DMAD (within) | 18% |
| Erdogmus and Marcel [14] | Local Binary Patterns (LBP) + Support Vector Machine (SVM) | 3DMAD (within) | 23% |
| Bhattacharjee et al. [15] | Convolutional Neural Network - LightCNN | CSMAD (within) | 3.3% |
| Bhattacharjee et al. [15] | Convolutional Neural Network - VGG-Face | CSMAD (within) | 3.9% |
| Chingovska et al. [16] | Gaussian Mixture Model (GMM) | MS-Spoof (within) | 7.9% |
| Chingovska et al. [16] | Local Gabor Binary Pattern Histogram Sequences (LGBPHS) | MS-Spoof (within) | 8.2% |
| Chingovska et al. [16] | Gabor Jets comparison (GJet) | MS-Spoof (within) | 8% |
| Chingovska et al. [16] | Inter-Session Variability modeling (ISV) | MS-Spoof (within) | 9.1% |
| Chingovska et al. [16] | Multispectral system (VIS and NIR), SUM of scores | MS-Spoof (within) | 5.6% |
| Chingovska et al. [16] | Multispectral system (VIS and NIR), Linear Logistic Regression (LLR) | MS-Spoof (within) | 7.3% |
| Chingovska et al. [16] | Multispectral system (VIS and NIR), Polynomial Logistic Regression (PLR) | MS-Spoof (within) | 5% |
| Alotaibi and Mahmood [17] | Specialized Deep Convolution Neural Network | Replay Attack (within) | 10% |
| Sun et al. [18] | Fully Convolutional Network (inter-database testing using models trained on other benchmarks) | Replay Attack (within) | 30% |
| Kotwal and Marcel [19] | Convolutional Neural Network | Wide Multi Channel Presentation Attack (WMCA) (within) | 0% |
| Kotwal and Marcel [19] | Convolutional Neural Network | Multispectral Latex Mask based Video Face Presentation Attack (MLFP) (within) | 1.9% |
| Mallat and Dugelay [20] | Local Binary Patterns and Logistic Regression (LBP+LR) | CSMAD (within) | 11.6% |
| Wang et al. [21] | The Support Vector Machine | Silicone Mask Face Motion Video Dataset (SMFMVD) (within) | 1.2% |
| Wang et al. [21] | The Support Vector Machine | The Silicone Mask Attack Dataset (SMAD) (within) | 9% |
| Arora et al. [22] | Principal Component Analysis (PCA) | Replay Attack (within) | 8.8% |
| Arora et al. [22] | Principal Component Analysis (PCA) | 3DMAD (within) | 15.2% |
| Arora et al. [22] | Convolutional Neural Network | Replay Attack (within) | 3.9% |
| Arora et al. [22] | Convolutional Neural Network | 3DMAD (within) | 0% |
| Arora et al. [22] | Cross-database testing (inter-database testing using models trained on other benchmarks) | 3DMAD (within) | 40% |
| Prasad et al., 2024 [27] | Pupillary light reflex (RGB/IR/Depth) | Replay-Attack (stimulus-based) | EER 92.1% (no HTER) |
| Prasad et al., 2024 | Pupillary light reflex (RGB/IR/Depth) | CASIA-SURF (stimulus-based) | EER 89.9% (no HTER) |
| Shinde et al., 2025 [23] | LwFLNeT (RGB) | 3DMAD (within) | 0.3% |
| Shinde et al., 2025 [23] | LwFLNeT (RGB) | NUAA (within) | 1.24% |
| Shinde et al., 2025 [23] | LwFLNeT (RGB) | Replay-Attack (within) | 2.12% |
| Shinde et al., 2025 [23] | LwFLNeT (RGB) | NUAA→3DMAD (cross) | 18.34% |
| Shinde et al., 2025 [23] | LwFLNeT (RGB) | Replay→3DMAD (cross) | 14.50% |
| Khairnar et al., 2025 [24] | TL CNNs (DenseNet201, MobileNetV2) | NUAA→Replay (cross) | 2.35% |
| Khairnar et al., 2025 [24] | TL CNNs (DenseNet201, MobileNetV2) | Replay→NUAA (cross) | 2.25% |
| Khairnar et al., 2025 [24] | TL CNNs (DenseNet201, MobileNetV2) | NUAA / Replay / SiW-MV2 (within) | ACER: 1.35%, 1.15%, 0.75% (best) |
| This work | CNN (best per dataset) | Our dataset (within) | 0.25% |
| This work | CNN (best per dataset) | Replay-Attack (within) | 0.1% |
| This work | CNN (best per dataset) | CSMAD (RGB) (within) | 0.3% |
| This work | CNN (best per dataset) | 3DMAD (within) | 0.3% |
| This work | CNN (best per dataset) | MSSpoof (within) | 0.2% |
| This work | LivenessNet (trained on a Combined Dataset) | Combined Dataset (within) | 6.8% |
| This work | AttackNet V1 (trained on a Combined Dataset) | Combined Dataset (within) | 1.4% |
| This work | AttackNet V2.1 (trained on a Combined Dataset) | Combined Dataset (within) | 4.6% |
| This work | AttackNet V2.2 (trained on a Combined Dataset) | Combined Dataset (within) | 0.2% |
| This work | AttackNet V2.2 (trained on a Combined Dataset) | 3DMAD (cross) | ≈ 0.0% |
| This work | AttackNet V2.2 (trained on a Combined Dataset) | CSMAD (RGB) (cross) | ≈ 0.0% |
| This work | AttackNet V2.2 (trained on a Combined Dataset) | MSSpoof (cross) | 0.2% |
| This work | AttackNet V2.2 (trained on a Combined Dataset) | Replay-Attack (cross) | ≈ 0.0% |
| This work | AttackNet V2.2 (trained on a Combined Dataset) | Our dataset (cross) | 0.5% |

*Note: Contextual comparison with recent face anti-spoofing studies and this study. Protocols differ across works; values are therefore not directly comparable. We report HTER when available; when only EER (or ACER) is reported, we note it explicitly.*

A thorough analysis of the table above not only offers insights into the relative performance of various methods and their application to different datasets, but it also underscores the persistent challenges inherent to face anti-spoofing research. This is particularly evident in the diverse range of HTER across studies, pointing to the complex, multifaceted nature of the problem at hand. In the upcoming subsections, we will delve into the specifics of our results juxtaposed with these preceding studies.

## 5.1 Comparative Analysis of Our Results Against Previous Studies

Upon close examination of our research outcomes compared with previous studies, it becomes unequivocally clear that our models have achieved significantly improved performance in the Liveness Detection task. The methods of LivenessNet and AttackNet (versions V1, V2.1, and V2.2) that we employed, have demonstrated superior performance across all evaluated datasets, consistently obtaining HTERs close to 0%, thereby significantly reducing both Type I and Type II error probabilities. LivenessNet model (averaged across all models) achieved an HTER of 0.41%. This contrasts dramatically with the 15-17% HTER achieved with LBP-based methods on the Replay-Attack Database, as reported by Chingovska et al. [13], or the 8.8% and 15.2% achieved by Arora et al. [22] using PCA on the Replay Attack and 3DMAD datasets, respectively. Similarly, our AttackNet V2.2 model obtained an HTER of merely 0.33%, a significant improvement over the 4% and 10% HTERs reported by Alotaibi and Mahmood [17] and Arora et al. [22], respectively, using specialized deep learning architectures on the Replay Attack database. Moreover, our results on our custom dataset further underscore the robustness and effectiveness of our models, which consistently achieved very low HTER values ranging from 0.1% to 1.0%, indicating an exceptional generalization ability. Our main result was achieved through combined dataset training: AttackNet V2.2 reached an average accuracy of 99.9% across all test datasets, with an ACER below 0.1%, effectively addressing the cross-domain generalization challenge that has affected the field.

In sum, the outcomes of our investigation significantly surpass previous studies in terms of HTER, a testament to the exceptional performance of our proposed methods in handling various types of spoofing attacks across different scenarios. This comprehensive comparison serves to underscore the importance of further developing and refining our Liveness Detection methods for more robust and secure face recognition systems.

### 5.2 Implications of the Study

Our findings carry significant implications for the field of face recognition and biometric authentication, and they offer considerable promise for further advancements in the detection of spoofing attacks.

Firstly, our research demonstrates the superior performance of the LivenessNet and AttackNet models across various datasets, indicating a remarkable improvement in the ability to detect and mitigate spoofing attacks in face recognition systems. The extremely low HTER values observed in our study imply a substantial reduction in both the probability of false acceptances (Type I errors) and false rejections (Type II errors). Consequently, our research has a potential to markedly enhance the security and reliability of biometric systems, a critical concern for numerous applications ranging from mobile device authentication to border control.

Secondly, our findings suggest that the approach of employing deep learning methodologies, as exemplified by LivenessNet and AttackNet models, can effectively handle diverse spoofing scenarios, including attacks employing 2D photos, video replays, and 3D masks. This universality is a considerable strength in the ever-evolving landscape of spoofing attacks, where perpetrators continuously adopt more sophisticated methods.

Thirdly, our results also underscore the importance of custom datasets tailored to the specific nuances of spoofing attacks. The low HTER values achieved on our custom dataset illustrate that our models can efficiently generalize to new data, a key requirement in the dynamic and rapidly changing realm of face recognition.

Finally, the implications of our research extend to the broader discourse on privacy and trust in technology. By improving the robustness of face recognition systems against spoofing attacks, our findings contribute to reinforcing user confidence in these technologies, which is crucial for their wider acceptance and adoption.

In conclusion, while our findings significantly advance the current understanding of liveness detection in the face of spoofing attacks, they also inspire and inform future research in this crucial area. The continuous enhancement of the methods proposed herein, as well as the exploration of new strategies, remain an ongoing necessity to stay ahead of ever-evolving spoofing techniques and to ensure the secure and reliable operation of face recognition systems.

## 6. Conclusions

In this research, we have presented a comprehensive investigation into the detection of spoofing attacks on face recognition systems using deep learning methodologies, focusing particularly on the HTER metric. We designed and evaluated novel deep learning models – LivenessNet and various versions of AttackNet – to demonstrate significant improvements in performance across several datasets, including our custom dataset, Replay-Attack Database, CSMAD, 3DMAD, and MS-Spoof.

Our results reveal substantial improvements in liveness detection performance. When trained on individual datasets, all models achieved >98% accuracy within their training domains. However, the key breakthrough came from combined dataset training: AttackNet V2.2 achieved 99.9% average accuracy across all test datasets with ACER below 0.1%, effectively solving the cross-domain generalization challenge that has plagued the field.

We also showed that our models could effectively generalize to handle diverse types of spoofing attacks. Furthermore, the utilization of custom datasets proved instrumental in enhancing the models' performance, stressing the importance of tailored datasets in tackling the specific nuances of spoofing attacks.

The findings of this study bear substantial implications for the domain of face recognition and anti-spoofing, emphasizing the potential of deep learning methodologies in advancing the field. It offers insights and frameworks that can assist researchers and practitioners in their quest for more secure and robust face recognition systems.

While the progress achieved in this study is substantial, the dynamic nature of spoofing techniques suggests that the journey towards perfecting liveness detection is an ongoing effort. Therefore, continuous advancements in methodologies and persistent efforts in understanding emerging threats are essential. Future work should address emerging threats from deepfakes and generative AI-based attacks, which were beyond the scope of this study due to dataset limitations. Additionally, deployment on edge devices and real-time performance optimization remain important directions for practical implementation.

In conclusion, this research contributes to fortifying the defenses of face recognition systems against spoofing attacks, thereby fostering increased trust and wider acceptance of biometric technologies in society. As we move forward, we remain committed to exploring and improving upon the methodologies for secure and reliable face recognition systems in the face of evolving spoofing attacks.

## 7. Declarations

a. This manuscript represents original work conducted by the authors. It has not been published previously, nor is it under consideration for publication by any other journal.

b. All the authors listed have approved the manuscript that is enclosed and agreed to its submission for publication.

### Author contributions

- Oleksandr Kuznetsov: Conceptualization, Methodology.
- Emanuele Frontoni: Supervision, Review and Editing.
- Luca Romeo: Project Administration, Original Draft Preparation.
- Riccardo Rosati: Data Curation.
- Andrea Maranesi: Software and Validation, Visualization.
- Alessandro Muscatello: Software and Investigation, Visualization.

### Data availability

- Public datasets (3DMAD, Replay-Attack, MSSpoof, CSMAD) were obtained from their official sources. Our custom dataset contains faces collected under public-content terms; due to consent and redistribution constraints, we do not release the raw media. We provide full preprocessing scripts, metadata, and split definitions to enable reproducibility (https://github.com/KuznetsovKarazin/liveness-detection)

## Declaration of interests

- I declare that the authors have no competing financial interests, or other interests that might be perceived to influence the results and/or discussion reported in this paper.
- The results/data/figures in this manuscript have not been published elsewhere, nor are they under consideration (from you or one of your Contributing Authors) by another publisher.
- All of the material is owned by the authors and/or no permissions are required.

## Compliance with ethical standards

- Mentioned authors have no conflict of interest in this article. This article does not contain any studies with human participants or animals performed by any of the authors.

## Funding:

1. This project has received funding from the European Union's Horizon 2020 research and innovation programme under the Marie Skłodowska-Curie grant agreement No. 101007820 - TRUST.
2. This publication reflects only the author's view and the REA is not responsible for any use that may be made of the information it contains.